\documentclass[runningheads]{llncs}
\usepackage{graphicx}
\usepackage{amsmath,amssymb,xspace}

\usepackage{color}
\usepackage{multirow}
\usepackage{pifont}
\usepackage{anyfontsize}
\usepackage{enumitem}
\usepackage[T1]{fontenc}

\usepackage{array}

\newcolumntype{L}{>{\centering\arraybackslash}m{3cm}}
\newcommand{\cfbox}[2]{{\color{#1} \fbox{#2}}}

\definecolor{olive}{rgb}{0.0, 0.5, 0.0}

\newcommand{\tabincell}[2]{\begin{tabular}{@{}#1@{}}#2/end{tabular}}

\newcommand{\xmark}{\ding{55}}

\newcommand{\std}[1]{\tiny{$\pm$#1}}

\usepackage{amsmath,amsfonts}
\usepackage{amssymb,amsopn}
\usepackage{bm} % bold symbol

\usepackage{amssymb}
\usepackage{amsmath,amsfonts}
\usepackage{amsopn}

\usepackage{bm} % bold symbol

\newcommand{\vct}[1]{\boldsymbol{#1}} % vector
\newcommand{\cst}[1]{\mathsf{#1}}  % constant

\newcommand{\ProbOpr}[1]{\mathbb{#1}}
\newcommand{\expect}[2]{%
\ifthenelse{\equal{#2}{}}{\ProbOpr{E}_{#1}}
{\ifthenelse{\equal{#1}{}}{\ProbOpr{E}\left[#2\right]}{\ProbOpr{E}_{#1}\left[#2\right]}}} % Expectation: syntax: E{1}{2} = E_1[2], E{}{2}=E[2], E{1}{} = E_1
\newcommand{\var}[2]{%
\ifthenelse{\equal{#2}{}}{\ProbOpr{VAR}_{#1}}
{\ifthenelse{\equal{#1}{}}{\ProbOpr{VAR}\left[#2\right]}{\ProbOpr{VAR}_{#1}\left[#2\right]}}} % Expectation: syntax: V{1}{2} = V_1[2], V{}{2}=V[2], V{1}{} = V_1
\newcommand{\vc}{\vct{c}}

\newcommand{\vp}{\vct{p}}
\newcommand{\vx}{{\vct{x}}}

\newcommand{\vv}{\vct{v}}
\newcommand{\vw}{\vct{w}}

\newcommand{\cn}{\cst{n}}

\newcommand{\cm}{\cst{m}}

\newcommand{\vs}{\vct{s}}

\newcommand{\vh}{\vct{h}}

\newcommand{\eat}[1]{}

\newcommand{\enc}{\textsc{enc}\xspace}
\newcommand{\dec}{\textsc{dec}\xspace}
\newcommand{\seq}{\textsc{seq2seq}\xspace}
\newcommand{\fse}{\textsc{fse}\xspace}
\newcommand{\hse}{\textsc{hse}\xspace}
\newcommand{\thse}{{\textsc{hse}\scriptsize{[$\tau$=0]}}\xspace}
\newcommand{\match}{\textsc{match}\xspace}

\newcommand{\alg}[1]{\textsc{#1}\xspace}

\newcommand{\ie}{\emph{i.e.}}
\newcommand{\eg}{\emph{e.g.}}

\begin{document}
\title{Cross-Modal and Hierarchical Modeling of Video and Text}

\titlerunning{Cross-Modal and Hierarchical Modeling of Video and Text}
\author{
Bowen Zhang\thanks{equal contribution}\inst{1}\and
Hexiang Hu$^\star$\inst{1}\and
Fei Sha\inst{2}}
\authorrunning{B. Zhang, H. Hu, and F. Sha}
\institute{Dept. of Computer Science, U. of Southern California, Los Angeles, CA 90089 \and
Netflix,  5808 Sunset Blvd, Los Angeles, CA 90028 \\
\email{zhan734@usc.edu,hexiangh@usc.edu, fsha@netflix.com\thanks{\it{On leave from U. of Southern California (feisha@usc.edu)}}}}
\maketitle
\begin{abstract}
% !TEX root = Arxiv_ECCV_BZ_HH_FS.tex
Visual data and text data are composed of information at multiple granularities. A video can describe a complex scene that is composed of multiple clips or shots, where each depicts a semantically coherent event or action.    Similarly, a  paragraph may contain sentences with different topics, which collectively conveys a coherent message or story. In this paper, we investigate the modeling techniques for such hierarchical sequential data where there are correspondences across multiple modalities. Specifically, we introduce hierarchical sequence embedding (\hse), a generic model for embedding sequential data of different modalities into hierarchically semantic spaces, with either explicit or implicit correspondence information. We perform empirical studies on large-scale video and paragraph retrieval datasets and demonstrated superior performance by the proposed methods. Furthermore, we examine the effectiveness of our learned embeddings when applied to downstream tasks. We show its utility in zero-shot action recognition and video captioning.
		

\end{abstract}

% !TEX root = Arxiv_ECCV_BZ_HH_FS.tex
\section{Introduction}
\label{sIntro}

Recently, there has been an intensive interest in multi-modal learning of vision + language. A few challenging tasks have been proposed: visual semantic embedding (VSE)~\cite{kiros2014unifying,kiela2014learning,collell2016image}, image captioning~\cite{vinyals2015show,xu2015show,karpathy2015deep,lin2014microsoft}, and visual question answering (VQA)~\cite{antol2015vqa,zhu2016visual7w,chao2017being}. To jointly understand these two modalities of data and make inference over them, the main intuition is that different types of data can share a common semantic representation space. Examples are embedding images and the visual categories~\cite{frome2013devise}, embedding images and texts for VSE~\cite{kiros2014unifying}, and embedding images, questions, and answers for VQA~\cite{hu2018learning}. Once embedded into this common (vector) space, similarity and distances among originally heterogeneous data can be captured by learning algorithms.

While there has been a rich study on how to discover this shared semantic representation on structures such as images, noun phrases (visual object or action categories) and sentences (such as captions, questions, answers), less is known about how to achieve so on more complex structures such as videos and paragraphs of texts~\footnote{We use paragraphs and documents interchangeably throughout this work.}. There are conceptual challenges: while complex structured data can be mapped to vector spaces (for instance, using deep architectures~\cite{krizhevsky2012imagenet,he2016deep}), it is not clear whether the intrinsic structures in those data's original format, after being transformed to the vectorial representations, still maintain their correspondence and relevance across modalities.

Take the dense video description task as an example~\cite{krishna2017dense}. The task is to describe a video which is made of short, coherent and meaningful clips. (Note that those clips could overlap temporally.) Due to its narrowly focused semantic content, each clip is then describable with a sentence. The description for the whole video is then a paragraph of texts with sentences linearly arranged in order. Arguably, a corresponding pair of video and its descriptive paragraph can be embedded into a semantic space where their embeddings are close to each other, using a vanilla learning model by ignoring the boundaries of clips and sentences and treating as a sequence of continually flowing visual frames and words.  However, for such a modeling strategy, it is opaque that if and how the correspondences at the ``lower level'' (\ie \xspace clips versus sentences) are useful in either deriving the embeddings or using the embeddings to perform downstream tasks such as video or text retrieval.

\begin{figure}[t]
\centering
\includegraphics[width=0.8\columnwidth]{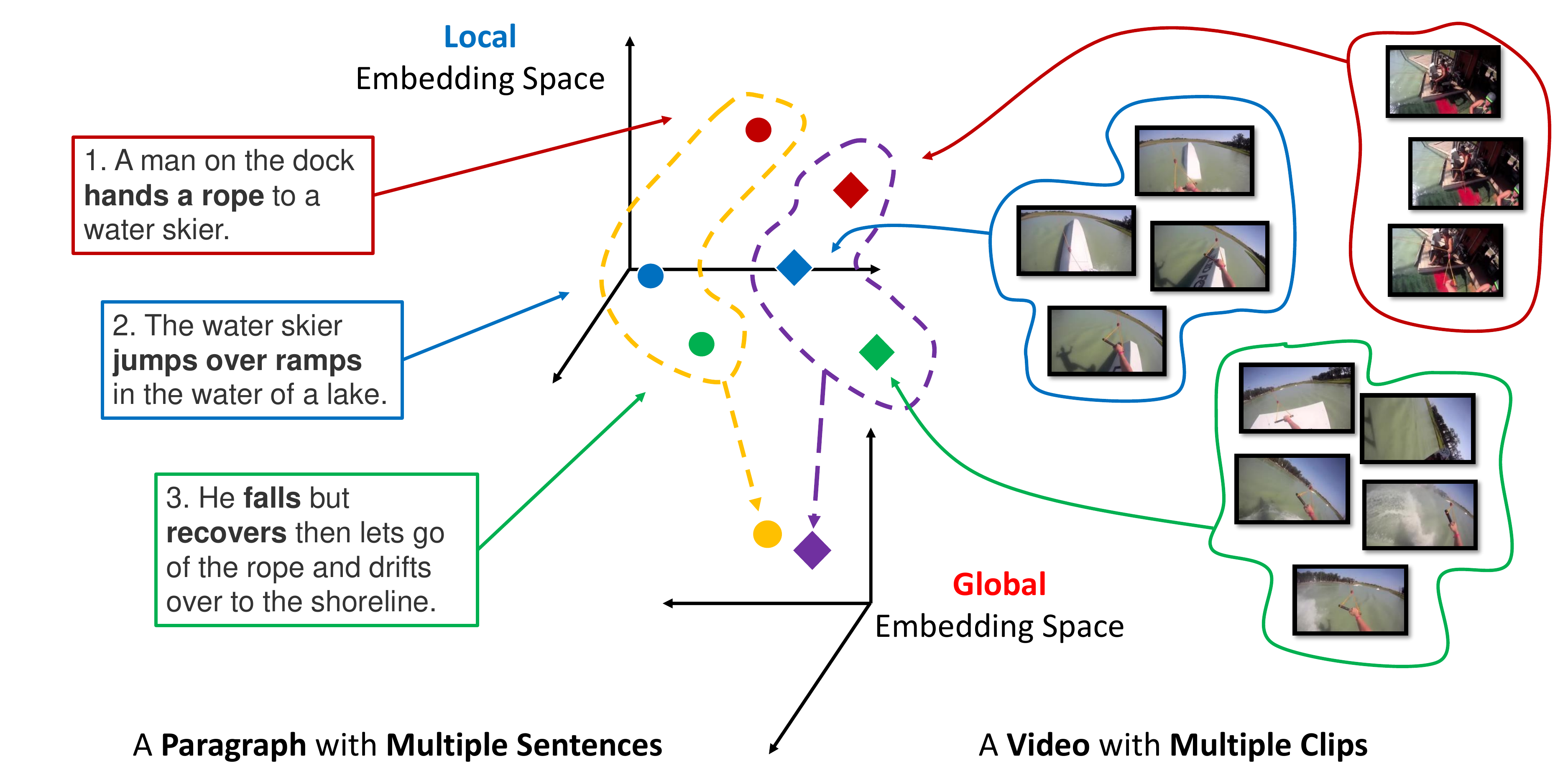}
\caption{{\small Conceptual diagram of our approach for cross-modal modeling of video and texts. The main idea is to embed both low-level (clips and sentences) and high-level (video and paragraph) in their own semantic spaces coherently. As shown in the figure, the 3 sentences (and the corresponding 3 clips) are mapped into a local embedding space where the corresponding pairs of clips and sentences are placed close to each other. As a whole, the videos and the paragraphs are mapped into a global semantic space where their embeddings are close. See Fig.~\ref{fHSE} and texts for details.}}
\label{fConcept}
\end{figure}

Addressing these deficiencies, we propose a novel cross-modal learning approach to model both videos and texts jointly. The main idea is schematically illustrated in Fig.~\ref{fConcept}. Our approach is mindful of the intrinsic hierarchical structures of both videos and texts, and models them with hierarchical sequence learning models such as GRUs~\cite{chung2014empirical}. However, as opposed to methods which disregard low-level correspondences, we exploit them by deriving loss functions to ensure the embeddings for the clips and sentences are also in accordance in their own (shared) semantic space. Those low-level embeddings in turn strengthen the desiderata that videos and paragraphs are embedded coherently.  We demonstrate the advantages of the proposed model in a range of tasks including video and text retrieval, zero-shot action recognition and video description.

The rest of the paper is organized as follows. In section~\ref{sRelated}, we discuss related work. We describe our proposed approach in section~\ref{sApproach}, followed by extensive experimental results and ablation studies in section~\ref{experiments}. We conclude in section~\ref{conclusion}.

% !TEX root = Arxiv_ECCV_BZ_HH_FS.tex
\section{Related Work}
\label{sRelated}

\paragraph{Hierarchical Sequence Embedding Models.}
Embedding images, videos, and textual data has been very popular with the rise of deep learning. The most related works to ours are ~\cite{li2015hierarchical} and ~\cite{pan2016hierarchical}. The former models the paragraph using a hierarchical auto-encoder for text modeling~\cite{li2015hierarchical}, and the later uses a hierarchical RNN for videos and a one-layer RNN for caption generation. In contrast, our work models both modalities hierarchically and learn the parameters by leveraging the correspondences across modalities. Works motivated by other application scenarios usually explore hierarchical modeling in one modality~\cite{niu2017hierarchical,yu2016video,zhang2016video}.

\paragraph{Cross-modal Embedding Learning.}
There has been a rich history to learn embeddings for images and smaller linguistic units (such as words and noun phrases). DeViSE~\cite{frome2013devise} learns to align the latent embeddings of visual data and names of the visual object categories.  ReViSE~\cite{tsai2017learning} uses auto-encoders to derive embeddings for images and words which allow them to leverage unlabeled data. In contrast to previous methods, our approach models both videos and texts hierarchically, bridging the embeddings at different granularities using discriminative loss computed on corresponded pairs (\ie \xspace videos vs. paragraphs).

\paragraph{Action Recognition in Videos.}
Deep learning has brought significant improvement to video understanding~\cite{simonyan2014two,tran2015learning,feichtenhofer2016convolutional,wang2016temporal,zhang2016real,wu2017compressed} on large-scale action recognition datasets~\cite{heilbron2015activitynet,soomro2012ucf101,kay2017kinetics} in the past decade. Most of them~\cite{simonyan2014two,feichtenhofer2016convolutional,wang2016temporal} employed deep convolutional neural network to learn appearance feature and motion information respectively. Based on the spatial-temporal feature from these video modeling methods, we learn video semantic embedding to match the holistic video representation to text representation. To evaluate the generalization of our learned video semantic representation, we evaluate the model directly on the challenging action recognition benchmark. (Details in Section~\ref{exp:action_recognition})

% !TEX root = Arxiv_ECCV_BZ_HH_FS.tex
\section{Approach}
\label{sApproach}

We begin by describing the problem settings and introducing necessary notations. We then describe the standard sequential modeling technique, ignoring the hierarchical structures in the data. Finally, we describe our approach.

\subsection{Settings and Notations}

We are interested in modeling videos and texts that are paired in correspondence In the later section,  we describe how to generalize this where there is no one to one correspondence.

A video $v$ has $\cn$ clips (or subshots), where each clip $c_i$ contains $\cn_i$ frames. Each frame is represented by a visual feature vector $\vx_{ij}$. This feature vector can be derived in many ways, for instance, by feeding the frame (and its contextual frames) to a convolution neural net and using the outputs from the penultimate layer.  Likewise, we assume there is a paragraph of texts describing the video. The paragraph $p$ contains $\cn$ sentences, one for each video clip. Let $s_i$ denote the $i$th sentence and $\vw_{ij}$ the feature for the $j$th word out of $\cn'_i$ words.  We denote by $\mathcal{D}= \{ (v_k, p_k)\}$ a set of corresponding videos and text descriptions.

We compute a clip vector embedding $\vc_i$ from the frame features $\{\vx_{ij}\}$, and a sentence embedding $\vs_i$ from the word features $\{\vw_{ij}\}$. From those, we derive $\vv$ and $\vp$, the embedding for the video and the paragraph, respectively.

\label{sFSM}
\begin{figure}[t]
	\centering
	\includegraphics[width=0.7\textwidth,trim={4cm 4cm 4cm 4cm},clip]{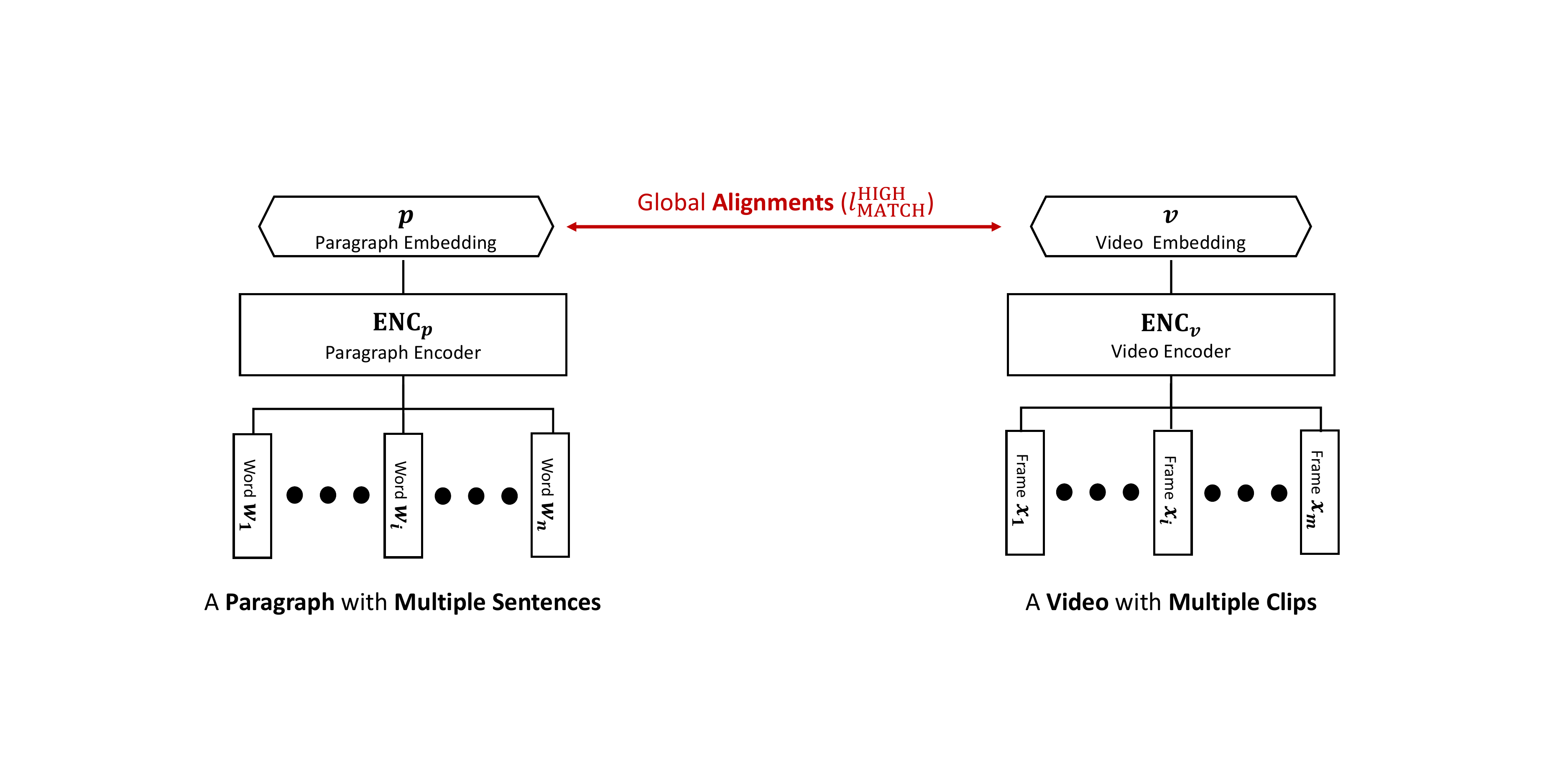}
	\caption{{\small Flat sequence modeling of videos and texts, ignoring the hierarchical structures in either and regarding the video (paragraph) as a sequence of frames (words). } }
	\label{fFSE}
\end{figure}

\subsection{Flat Sequence Modeling}

Many sequence-to-sequence (\seq) methods leverage the encoder-decoder structure~\cite{sutskever2014sequence,luong2015effective} to model the process of transforming from the input sequence to the output sequence. In particular, the encoder, which is composed of a layer of long short-term memory units (LSTMs)~\cite{hochreiter1997long} or Gated Recurrent Units (GRUs)~\cite{chung2014empirical}, transforms the input sequence into a vector as the embedding $\vh$. The similarly constructed decoder takes $\vh$ as input and outputs another sequence.

The original \seq methods do not consider the hierarchical structures in videos or texts. We refer the embeddings as \emph{flat sequence embedding} (\fse):
\begin{equation}
\vv = \enc_v ( \{\vx_{ij}\}), \ \  \vp = \enc_p ( \{\vw_{ij}\}),
\end{equation}

Fig.~\ref{fFSE} schematically illustrates this idea. We measure how well the videos and the texts are aligned by the following cosine similarity
\begin{equation}
\match(v, p) = \vv^\top \vp /{\lVert \vv \rVert \lVert \vp \rVert }
\end{equation}

\begin{figure}[t]
	\centering
	\includegraphics[width=0.99\textwidth,trim={0.8cm 0 0.8cm 0.5cm},clip]{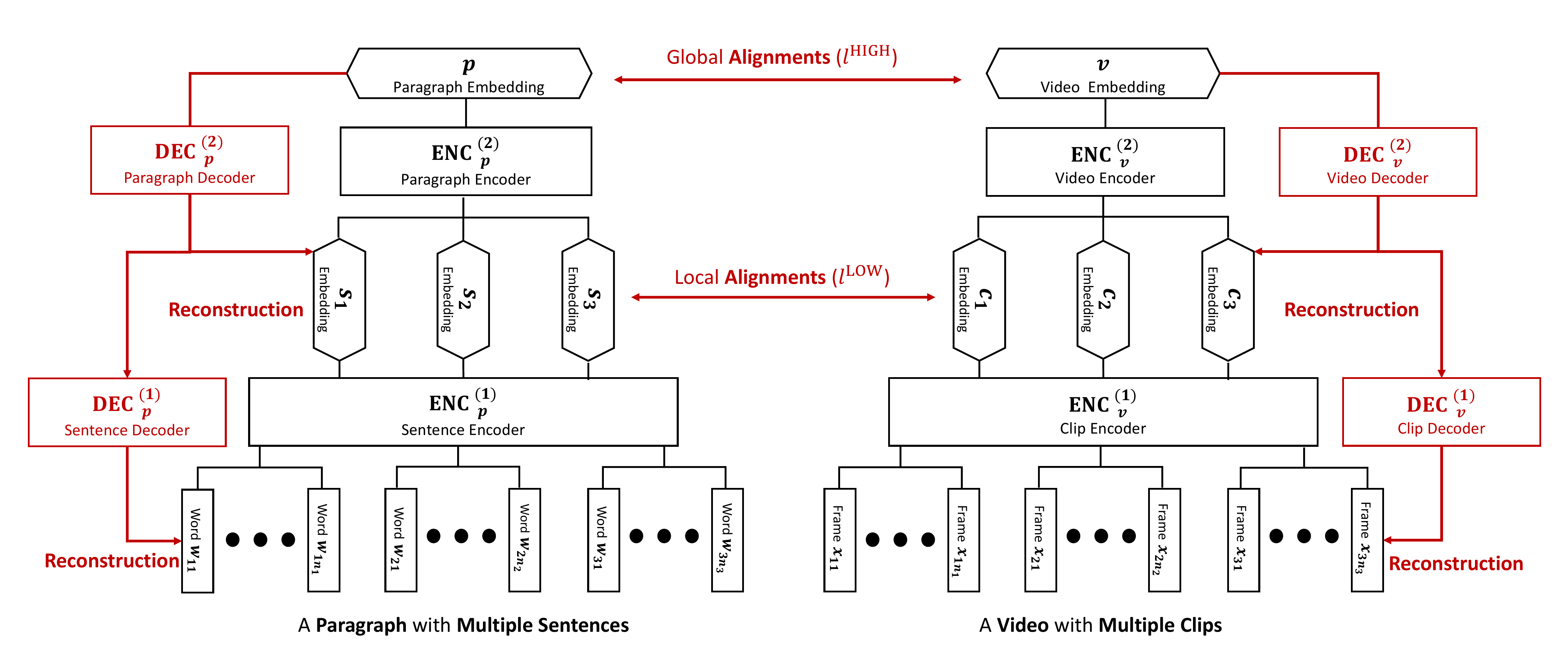} \\
    \caption{{\small Hierarchical cross-modal modeling of videos and texts. We differ from previous works~\cite{li2015hierarchical,pan2016hierarchical} in two aspects (components in red color): layer-wise reconstruction through decoders, and matching at both global and local levels.  See texts for details.}}
\label{fHSE}
\end{figure}

\subsection{Hierarchical Sequence Modeling}
\label{sHSM}

One drawback of flat sequential modeling is that the LSTM/GRU layer needs to have a sufficient number of units to model well the potential long-range dependency among video frames (or words). This often complicates learning as the optimization becomes difficult~\cite{pascanu2013difficulty}.

We leverage the hierarchical structures in those data to overcome this deficiency: a video is made of clips which are  made of frames. In parallel, a paragraph of texts is made of sentences which in turn are made of words. Similar ideas have been explored in \cite{pan2016hierarchical,li2015hierarchical} and other previous works. The basic idea is illustrated in Fig.~\ref{fHSE}, where we also add components in red color to highlight our extensions.

\noindent\textbf{Hierarchical Sequence Embedding.} Given the hierarchical structures in Fig.~\ref{fHSE}, we can compute the embeddings using the forward paths
\begin{equation}
\begin{array}{ccc}
\vc_i = \enc_v^{(1)} ( \{\vx_{ij}, j=1, 2, \cn_i\}), & &\vv = \enc_v^{(2)} (\{\vc_i\})\\
\vs_i = \enc_p^{(1)} ( \{\vw_{ij},j=1, 2, \cn_i'\}), & &\vp = \enc_p^{(2)} (\{\vs_i\})
\end{array}
\end{equation}

\noindent\textbf{Learning with Discriminative Loss.} For videos and texts have strong correspondences where clips and sentences are paired, we optimize the encoders such that videos and texts are matched. To this end, we define two loss functions, corresponding to the matching at the low-level and the high-level respectively:
\begin{align}
 \ell_{\textsc{match}}^{\textsc{high}}  = \sum_k \sum_{k'\ne k} [\alpha & + \match(\vv_k, \vp_k) - \match(\vv_{k'}, \vp_k)]_+ \nonumber\\
&   +  [\alpha+ \match(\vv_k, \vp_k) - \match(\vv_k, \vp_{k'})]_+\\
 \ell_{\textsc{match}}^{\textsc{low}}  = \sum_k \sum_i \sum_{(k',i')\ne (k,i)}   [\beta &+ \match(\vc_{ki}, \vs_{ki}) - \match(\vc_{k'i'}, \vs_{ki})]_+  \nonumber \\
+ [\beta &+ \match(\vc_{ki}, \vs_{ki}) - \match(\vc_{ki}, \vs_{k'i'})]_+
\end{align}
These losses are margin-based losses~\cite{schroff2015facenet} where $\alpha$ and $\beta$ are positive numbers as the margins to separate matched pairs from unmatched ones. The function $[\cdot]_+$ is the standard hinge loss function.

\noindent\textbf{Learning with Contrastive Loss.} Assuming videos and texts are well clustered, we use the following loss to model their clustering in their own space.

\begin{align}
\ell_{\textsc{cluster}}^{\textsc{high}} = \sum_k \sum_{k'\ne k} [\gamma  + 1 - \match(\vv_{k'}, \vv_k)]_+  &+ [\gamma + 1 - \match(\vp_{k'}, \vp_k)]_+  \\
\ell_{\textsc{cluster}}^{\textsc{low}}  = \sum_k \sum_i \sum_{(k',i')\ne (k,i)}  [\eta  &+  1 - \match(\vc_{k'i'}, \vc_{ki})]_+ \nonumber \\
 + [\eta + 1 - &\match(\vs_{k'i'}, \vs_{ki})]_+
\end{align}
Note that the self-matching values $\match(\vv_k, \vv_k)$ and $\match(\vp_k, \vp_k)$  are 1 by definition. This loss can be computed on videos and texts alone and does not require them being matched.

\noindent\textbf{Learning with Unsupervised Layer-wise Reconstruction Loss.} Thus far, the matching loss focuses on matching across modality. The clustering loss focuses on separating between video/text data so that they do not overlap. None of them, however, focuses on the \emph{quality} of the modeling data itself. In what follows, we propose a layer-wise reconstruction loss -- when minimized, this loss ensures the learned video/text embedding faithfully preserves information in the data.

We first introduce a set of layer-wise decoders for both videos and texts. The key idea is to pair the encoders with decoders so that each pair of functions is an auto-encoder. Specifically, the decoder is also a layer of LSTM/GRU units, generating sequences of data. Thus, at the level of video (or paragraph), we will have a decoder to generate clips (or sentences). And at the level of clips (or sentences), we will have a decoder to generate frames (or words).  Concretely, we would like to minimize the difference between what are generated by the decoders and what are computed by encoders on the data. Let

\begin{equation}
\{\hat{\vc}_i\} = \dec_v^{(2)}(\vv),  \{\hat{\vs}_i\} =  \dec_p^{(2)}(\vp)
\end{equation}
be the two (high-level) decoders for videos and texts respectively. And similarly, for the decoder at the low-level
\begin{equation}
\{\hat{\vx}_{ij}\} = \dec_v^{(1)}(\hat{\vc}_i), \{\hat{\vw}_{ij}\} = \dec_p^{(1)}(\hat{\vs}_i)
\end{equation}

where the low-level decoders take each \emph{generated} clip and sentence embeddings as inputs and output sequences of generated frame and word embeddings.

\begin{align}
\label{eReconstruct}
\ell_{\textsc{reconstruct}} (v, p) & =  \sum_i \{\| \hat{\vc}_i- \vc_i\|_2^2 + \frac{1}{\cn_i}\sum_j \| \hat{\vx}_{ij}- \vx_{ij}\|_2^2\}\nonumber \\
& +  \sum_i \{\| \hat{\vs}_i- \vs_i\|_2^2 + \frac{1}{\cn_i'}\sum_j \| \hat{\vw}_{ij}- \vw_{ij}\|_2^2\}
\end{align}

Using those generated embeddings, we can construct a loss function characterizing how well the encoders encode the data pair $(v, p)$ (see Eq~\ref{eReconstruct}).

\subsection{Final Learning Objective and Its Extensions}

The final learning objective is to balance all those loss quantities

\begin{equation}
\ell =  \ell^{\textsc{high}} + \ell^{\textsc{low}} + \tau \sum_k \ell_{\textsc{reconstruct}} (\vv_k, \vp_k)
\end{equation}

where the high-level and low-level losses are defined as

\begin{align}
\ell^{\textsc{high}} =   \ell_{\textsc{match}}^{\textsc{high}} + \ell_{\textsc{cluster}}^{\textsc{high}}, \ \ \ell^{\textsc{low}} =   \ell_{\textsc{match}}^{\textsc{low}} + \ell_{\textsc{cluster}}^{\textsc{low}}
\end{align}

In our experiments, we will study the contribution by each term.

\noindent\textbf{Learning under Weak Correspondences.} Our idea can be also extended to the common setting where only high-level alignments are available. In fact, high-level coarse alignments of data are easier and more economical to obtain, compared to fine-grained alignments between each sub-level sentence and video clip.

Since we do not have enough information to define the low-level matching loss $\ell_{\textsc{match}}^{\textsc{low}}$ exactly, we resort to approximation. We first define an averaged matching over all pairs of clips and sentences for a pair of video and paragraph

\begin{equation}
\overline{\match}(v, p) = \frac{1}{\cn\cm} \sum_{c_i}\,\sum_{s_j} \match(\vc_i, \vs_j)
\end{equation}

where we relax the assumption that there is precisely the same number of sentences and clips. We use this averaged quantity to approximate the low-level matching loss

\begin{align}
\tilde{\ell}_{\textsc{match}}^{\textsc{low}} =  \sum_k \sum_{k'\ne k} [\beta'  &+ \overline{\match}(\vv_k, \vp_k) - \overline{\match}(\vv_{k'}, \vp_k)]_+ \nonumber \\
&  +  [\beta'+ \overline{\match}(\vv_k, \vp_k) - \overline{\match}(\vv_k, \vp_{k'})]_+
\end{align}

This objective will push a clip embedding closer to the embeddings of the sentences belonging to the corresponding video (and vice versa for sentences to the corresponding video). A more refined approximation involving a soft assignment of matching can also be derived, which will be left for future work.

% !TEX root = Arxiv_ECCV_BZ_HH_FS.tex
\section{Experiments}
\label{experiments}

We evaluate and demonstrate the advantage of  learning hierarchical cross-modal embedding with our proposed approach on several tasks:  (i) large-scale video-paragraph retrieval (Section~\ref{exp:retrieval}), (ii)  down-stream tasks such as video captioning (Section~\ref{exp:caption}), and (iii) action recognition (Section~\ref{exp:action_recognition}).

\subsection{Experiment Setups}
\label{exp:dataset}

\subsubsection{Datasets.} We evaluate on three large-scale video datasets:

\noindent\texttt{(1) ActivityNet Dense Caption~\cite{krishna2017dense}.} This variant of ActivityNet contains densely labeled temporal segments for 10,009 training and 4,917/4,885 (val1/val2) validation videos. Each video contains multiple clips and a corresponding paragraph with sentences aligned to the clips. In all our retrieval experiments, we follow the setting in~\cite{krishna2017dense} and report retrieval metrics such as recall@k (k=1,5,50) and median rank (MR). Following~\cite{krishna2017dense} we use ground-truth clip proposals as input for our main results. In addition, we also study our algorithm with a heuristic proposal method (see Section~\ref{exp:nogt}). In the main text, we report all results on validation set 1 (val1). Please refer to the Supp. Material for the results on val2. For video caption experiment, we follow \cite{krishna2017dense} and evaluate on the validation set (val1 and val2). Instead of using action proposal method, ground-truth video segmentation is used for training and evaluation. Performances are reported in Bleu@K, METEOR and CIDEr.

\noindent\texttt{(2) DiDeMo~\cite{anne2017localizing}.} The original goal of DiDeMo dataset is to locate the temporal segments that correspond to unambiguous natural language descriptions in a video. We re-purpose it for the task of video and paragraph retrieval. It contains 10,464 videos, 26,892 video clips and 40,543 sentences. The training, validation and testing split contain 8,395, 1,065  and 1,004 videos and corresponding paragraphs, respectively. Each video clip may correspond to one or many sentences. For the video and paragraph retrieval task, paragraphs are constructed by concatenating all sentences that corresponding to one video. Similar to the setting in ActivityNet, we use the ground-truth clip proposals as input.

\noindent\texttt{(3) ActivityNet Action Recognition~\cite{heilbron2015activitynet}.} We use ActivityNet V1.3 for aforementioned off-the-shelf action recognition. The dataset contains 14,950 untrimmed videos with 200 action classes, which is split into training and validation set. Training and validation set have 10,024 and 4,926 videos, respectively. Among all 200 action classes, 189 of the action classes have been covered by the vocabulary extracted from the paragraph corpus and 11 of the classes are unseen.

\subsubsection{Baselines and Our Methods.}
We use the \fse method (as described in Section~\ref{sFSM}) as a baseline model. It ignores the clip and sentence structures in the videos and paragraphs. We train a one-layer GRU directly on the extracted frame/word features and take their outputs as the embedding representing each modality. Results with C3D features are also included (see Table \ref{tab:anet:benchmark}).

Our method has two variants: when $\tau=0$, the method (\thse) simplifies to a stacked/hierarchical sequence models as used in~\cite{li2015hierarchical,pan2016hierarchical} except that they do not consider cross-modal learning with cross-modal matching loss while we do. We consider this as a very strong baseline. When $\tau\ne0$, the \hse takes full advantage of layer-wise reconstruction with multiple decoders, at different levels of the hierarchy.  In our experiments, this method gives the best results.

\subsubsection{Implementation Details.}
\label{exp:implement}
Following the settings of~\cite{krishna2017dense}, we extract the C3D features~\cite{tran2015learning} pretrained on Sports-1M dataset~\cite{karpathy2014large} for raw videos in ActivityNet. PCA is then used to reduce the dimensionality of the feature to 500.
To verify the generalization of our model across different sets of visual feature, as well as leveraging the state-of-the-art video models, we also employed recently proposed TSN-Inception V3 network~\cite{wang2016temporal} pre-trained on Kinetics~\cite{kay2017kinetics} dataset to extract visual features. Similarly, we extract TSN-Inception V3 feature for videos in Didemo dataset. We do not fine-tuning the convolutional neural network on the video along the training to reduce the computational cost. For word embedding, we use 300 dimension GloVe~\cite{pennington2014glove} features pre-trained on 840B common web-crawls. In all our experiments, we use GRU as sequence encoders. For \hse, we choose $\tau = 0.0005$ from tuning this hyper-parameter on the val2 set of ActivityNet retrieval dataset. The same $\tau$ value is used for experiments on DiDeMo, without further tuning. (More details in the Supp. Material)

\subsection{Results on Video-Paragraph Retrieval}
\label{exp:retrieval}

In this section, we first compare our proposed approach to the state-of-the-art algorithms, and then perform ablation studies on variants of our method, to evaluate the proposed learning objectives.

\begin{table}[t]
	\centering
	\small
	\caption{Video paragraph retrieval on ActivityNet (val1). Standard deviation from 3 random seeded experiments are also reported.}
	\begin{tabular}{ccccccccc}
		& \multicolumn{4}{c}{\text{ \textbf{P}aragraph } $\Rightarrow$ \text{ \textbf{V}ideo }} & \multicolumn{4}{c}{\text{ \textbf{V}ideo } $\Rightarrow$ \text{ \textbf{P}aragraph }} \\

		& R@1 & R@5 & R@50 & MR & R@1 & R@5 & R@50 & MR \\ \hline
				\multicolumn{9}{c}{C3D Feature with Dimensionality Reduction~\cite{tran2015learning}} \\ \hline

		\alg{lstm-yt}\cite{venugopalan2015sequence}         & 0.0 & 4.0 & 24.0 & 102.0 & 0.0 & 7.0 & 38.0 & 98.0 \\
		\alg{no context}\cite{venugopalan2014translating}   & 5.0 & 14.0 & 32.0 & 78.0 & 7.0 & 18.0 & 45.0 & 56.0 \\
		\alg{dense} online\cite{krishna2017dense}           & 10.0 & 32.0 & 60.0 & 36.0 & 17.0 & 34.0 & 70.0 & 33.0 \\
		\alg{dense} full\cite{krishna2017dense}             & 14.0 & 32.0 & 65.0 & 34.0 & 18.0 & 36.0 & 74.0 & 32.0 \\ \hline
		\fse          & 12.6\std{0.4} & 33.2\std{0.3}  & 77.6\std{0.3} & 12.0 & 11.5\std{0.5} & 31.8\std{0.3} & 77.7\std{0.3} & 13.0  \\
		\thse         & 32.8\std{0.3} & 62.3\std{0.4} & 90.5\std{0.1} & 3.0 & 32.0\std{0.6} & 62.5\std{0.5} & 90.5\std{0.3} & 3.0 \\
		\hse{\scriptsize[$\tau$=5e-4]} & 32.7\std{0.7} & 63.2\std{0.4} & 90.8\std{0.2} & 3.0 & 32.8\std{0.4} & 63.2\std{0.2} & 91.2\std{0.3} & 3.0 \\ \hline

		\multicolumn{9}{c}{Inception-V3 pre-trained on Kinetics~\cite{wang2017temporal}} \\ \hline
		\fse          & 18.2\std{0.2} &  44.8\std{0.4} & 89.1\std{0.3} & 7.0 & 16.7\std{0.8}  & 43.1\std{1.1}  & 88.4\std{0.3}  & 7.3  \\
		\thse         & 43.9\std{0.6} & 75.8\std{0.2} & 96.9\std{0.3} & 2.0 & 43.3\std{0.6} & 75.3\std{0.6} & 96.6\std{0.2} & 2.0 \\
		\hse{\scriptsize[$\tau$=5e-4]} & \textbf{44.4\std{0.5}} & \textbf{76.7\std{0.3}} & \textbf{97.1\std{0.1}} & \textbf{2.0} & \textbf{44.2\std{0.6}} & \textbf{76.7\std{0.3}} & \textbf{97.0\std{0.3}} & \textbf{2.0} \\
	\end{tabular}
	\label{tab:anet:benchmark}
\end{table}

\begin{table}[t]
	\centering
	\small
	\setlength\tabcolsep{3pt}
	\caption{Video paragraph retrieval on DiDeMo dataset. \alg{s2vt} method is re-implemented for retrieval task. }
	\begin{tabular}{c @{\quad} cccccccc}
		& \multicolumn{4}{c}{\text{ \textbf{P}aragraph } $\Rightarrow$ \text{ \textbf{V}ideo }} & \multicolumn{4}{c}{\text{ \textbf{V}ideo } $\Rightarrow$ \text{ \textbf{P}aragraph }} \\
		 & R@1 & R@5 & R@50 & MR & R@1 & R@5 & R@50 & MR \\ \hline
		 \alg{s2vt}\cite{venugopalan2014translating} & 11.9 & 33.6 & 76.5 & 13.0 & 13.2 & 33.6 & 76.5 & 15.0 \\ \hline
		\fse & 13.9\std{0.7}  & 36.0\std{0.8}  & 78.9\std{1.6}  & 11.0 &  13.1\std{0.5} & 33.9\std{0.4}  & 78.0\std{0.8}  & 12.0    \\
		\thse & \textbf{30.2\std{0.8}} & \textbf{60.5\std{1.1}} & 91.8\std{0.7} & 3.3 & 29.4\std{0.4} & 58.9\std{0.7} & 91.9\std{0.6} & 3.7\\
		\hse{\scriptsize[$\tau$=5e-4]} & 29.7\std{0.2} & 60.3\std{0.9} & \textbf{92.4\std{0.3}} & \textbf{3.3} & \textbf{30.1\std{1.2}} & \textbf{59.2\std{0.9}} & \textbf{92.1\std{0.5}} & \textbf{3.0}\\
	\end{tabular}
	\label{tab:didemo:benchmark}
\end{table}

\subsubsection{Main Results.}
\label{exp:sota}
We reported our results on ActivityNet Dense Caption val1 set and DiDeMo test set as Table~\ref{tab:anet:benchmark} and Table~\ref{tab:didemo:benchmark}, respectively. For both C3D and Inception V3 feature, we observed performances on our hierarchical models improved the previous state-of-the-art result by a large margin (on Recall@1, over $\sim 15\%$ improvement with C3D and $\sim 30\%$ improvement with InceptionV3). \alg{dense} full~\cite{krishna2017dense}, which models the flat sequences of clips, outperforms our \fse baseline as they augment each segment embedding with a weighted aggregated context embedding. However, it fails to model more complex temporal structures of video and paragraph, which leads to inferior performance to our \hse models.

Comparing to our flat baseline model, both \thse and \hse{\scriptsize[$\tau$=5e-4]} improve performances over all metrics in retrieval. It implies that hierarchical modeling can effectively capture the structure information and relationships over clips and sentences among videos and paragraphs. Moreover, we observe that \hse{\scriptsize[$\tau$=5e-4]} consistently improves over \thse across most retrieval metrics on both datasets. This attributes the importance of our layer-wise reconstruction objectives, which suggests that better generalization performances.
\begin{table}[t]
	\centering
	\small
	\caption{Ablation studies on the learning objectives. }
	\begin{tabular}{c|c@{\quad}ccccccc}
		& & & \multicolumn{3}{c}{\text{ \textbf{P}aragraph } $\Rightarrow$ \text{ \textbf{V}ideo }} & \multicolumn{3}{c}{\text{ \textbf{V}ideo } $\Rightarrow$ \text{ \textbf{P}aragraph }} \\
		Dataset &  & $\ell^{\textsc{low}}$ & R@1 & R@5 & R@50 & R@1 & R@5 & R@50 \\ \hline
		\multirow{6}{*}{\textbf{ActivityNet}} &
		\multirow{3}{*}{\thse}  & \xmark & 41.8\std{0.4} & 74.1\std{0.6} & 96.6\std{0.1} & 40.5\std{0.4} &  73.9\std{0.6} & 96.3\std{0.1}  \\
		&  & \alg{weak} & 42.6\std{0.4} & 74.8\std{0.3} & 96.7\std{0.1} & 41.3\std{0.2} & 74.7\std{0.4} & 96.5\std{0.1}  \\
		&  & \alg{strong} & 43.9\std{0.6} & 75.8\std{0.2} & 96.9\std{0.3} & 43.3\std{0.6} & 75.3\std{0.6} & 96.6\std{0.2} \\ \cline{2-9}
		& \multirow{3}{*}{\hse{\scriptsize[$\tau$=5e-4]}} & \xmark & 42.5\std{0.3} & 74.8\std{0.1} & 96.9\std{0.0} & 41.6\std{0.2} & 74.7\std{0.6} &96.6\std{0.1} \\
		&  & \alg{weak} & 43.0\std{0.6} & 75.2\std{0.4} & 96.9\std{0.1} & 41.5\std{0.1} & 75.2\std{0.6} & 96.8\std{0.2}  \\
		&  & \alg{strong} & 44.4\std{0.5} & 76.7\std{0.3} & 97.1\std{0.1} & 44.2\std{0.6} & 76.7\std{0.3} & 97.0\std{0.3} \\ \hline
		\multirow{6}{*}{\textbf{DiDeMo}} &
		\multirow{3}{*}{\thse} & \xmark & 27.1\std{1.9} & 59.1\std{0.4} & 92.2\std{0.3} & 27.3\std{1.0} & 57.6\std{0.5} & 91.3\std{1.2} \\
		&  & \alg{weak}  & 28.0\std{0.8} & 58.9\std{0.5} & 91.4\std{0.6} & 28.3\std{0.3} & 58.5\std{0.6} & 91.2\std{0.3} \\
		&  & \alg{strong} & 30.2\std{0.8} & 60.5\std{1.1} & 91.8\std{0.7} & 29.4\std{0.4} & 58.9\std{0.7} & 91.9\std{0.6} \\ \cline{2-9}
		& \multirow{3}{*}{\hse{\scriptsize[$\tau$=5e-4]}} & \xmark & 28.1\std{0.8} & 59.5\std{1.1} & 91.7\std{0.7} & 28.2\std{0.8} & 58.1\std{0.5} & 90.9\std{0.5} \\
		& & \alg{weak} & 28.7\std{2.1} & 59.1\std{0.2} & 91.6\std{0.7} & 28.3\std{0.8} & 59.2\std{0.6} & 91.1\std{0.1} \\
		& & \alg{strong} & {29.7\std{0.2}} & {60.3\std{0.9}} & 92.4\std{0.3} & {30.1\std{1.2}} & {59.2\std{0.9}} & 92.1\std{0.5} \\
	\end{tabular}
	\label{tab:ablation}
\end{table}

\subsubsection{Low-level Loss is Beneficial.}
\label{exp:ablation}
Table~\ref{tab:anet:benchmark} and Table~\ref{tab:didemo:benchmark} have shown results with optimizing both low-level and high-level objectives. In Table~\ref{tab:ablation}, we further performed ablation studies on the learning objectives. Note that rows with~\xmark~represent learning without low-level loss $\ell^{\textsc{low}}$. In all scenarios, joint learning with both low-level and high-level correspondences improves the retrieval performance.

\subsubsection{Learning with Weak Correspondences at Low-level.}
\label{exp:weak}
As mentioned in Section~\ref{sApproach}, our method can be extended to learn the low-level embedding with weak correspondence. We evaluate its effectiveness on  both ActivityNet and DiDeMo datasets. Performance are listed in Table~\ref{tab:ablation}. Note that for the rows of ``weak'',  no auxiliary alignments between sentences and clips are available during training.

Clearly, including low-level loss with weak correspondence (ie, correspondence only at the high-level) obtained superior performances when compared to models that do not include low-level loss at all. On several occasions, it even attains the same competitive result as including low-level loss with strong correspondences at the clip/sentence levels.

\begin{table}[t]
	\centering
	\small
	\setlength\tabcolsep{3pt}
	\caption{Performance of using proposal instead of ground truth on ActivityNet dataset}
	\begin{tabular}{cccccccc}
		& & \multicolumn{2}{c}{\text{ \textbf{P} } $\Rightarrow$ \text{ \textbf{V} }} & \multicolumn{2}{c}{\text{ \textbf{V} } $\Rightarrow$ \text{ \textbf{P} }} & & \\
		Proposal Method & \# Segments & R@1 & R@5 & R@1 & R@5 & Precision & Recall \\ \hline
	    {\hse + \alg{ssn} } & -
		& 10.4 & 31.9 & 10.8 & 31.7  & 1.5 & 17.1  \\
         \hline
		\multirow{4}{*}{\hse + \alg{uniform}}
		& 1  &  18.0 & 45.5  &  16.5 & 44.9  & 63.2 & 31.1 \\
		& 2  &  20.0 & 48.9  &  18.4 & 47.6  & 61.8 & 46.0  \\
		& 3  &  20.0 & 48.6  &  18.2 & 47.9  & 55.3 & 50.6  \\
		& 4  & 20.5 & 49.3 & 18.7 & 48.1  & 43.2 & 45.5  \\
         \hline
	    {\hse + \alg{ground truth}} & -
		& 44.4 & 76.7 & 44.2 & 76.7 & 100.0 & 100.0  \\
		 \hline
	    {\fse} & -
		& 18.2 & 44.8 & 16.7 & 43.1 & - & -  \\
	\end{tabular}
	\label{tab:proposal}
\end{table}

\subsubsection{Learning with Video Proposal Methods.}
\label{exp:nogt}
As using ground-truth temporal segments of videos is not a natural assumption, we perform experiments to validate the effectiveness of our method with proposal methods. Specifically, we experiment with two different proposal approaches: SSN~\cite{zhao2017temporal} pre-trained on ActivityNet action proposal and a heuristic uniform proposal. For uniform proposal of K segments, we meant naturally segmenting a video into K non-overlapping and equal-length temporal segments.

The results are summarized in Table~\ref{tab:proposal} (with columns of precision and recall being the performance metrics of the proposal methods). There are two main conclusions from these results: (1) The segments of Dense Caption dataset deviate significantly from the action proposals, therefore a pre-trained action proposal algorithm performs poorly. (2) Even with heuristic proposal methods, the performance of \hse is mostly better than (or comparable with) \fse. We leave to future work on identifying stronger methods for proposals.

\subsubsection{Retrieval with Incomplete Video and Paragraph.}
\label{exp:partial}
\begin{figure}[t]
	\centering
	\begin{tabular}{cc}
		\includegraphics[width=0.485\textwidth]{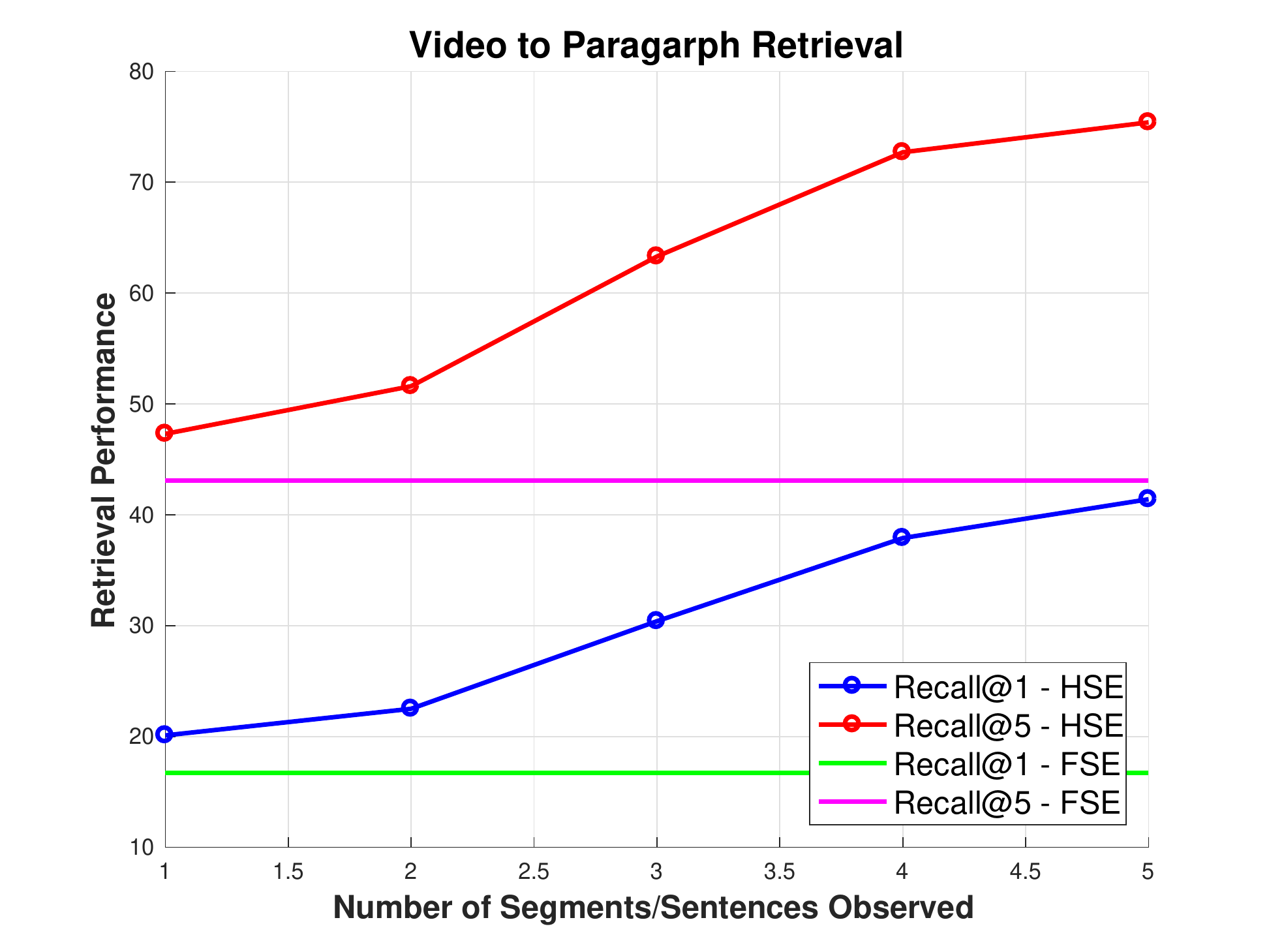} & \includegraphics[width=0.485\textwidth]{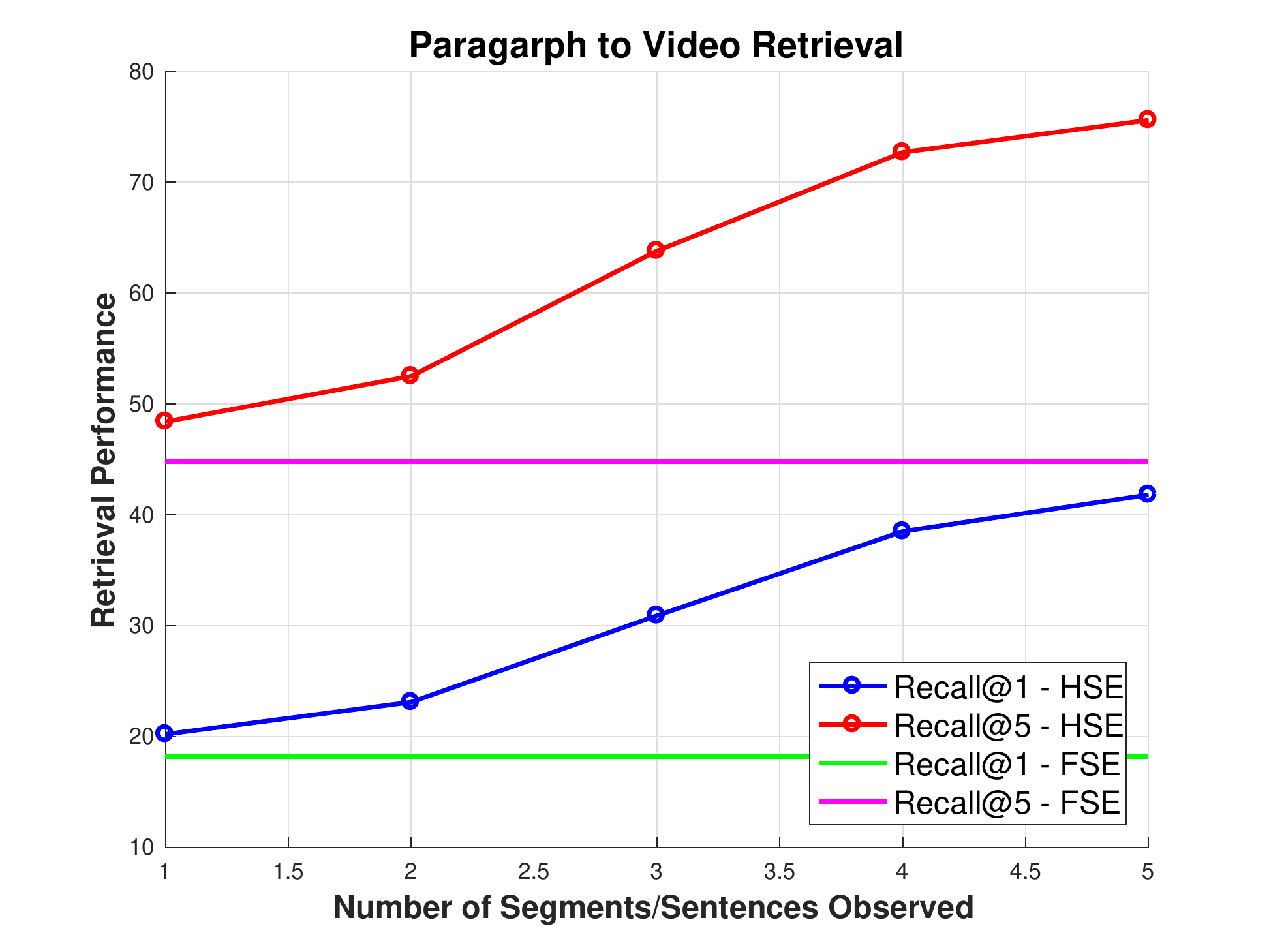}
	\end{tabular}
	\caption{Retrieval performance improves given more observed clips/sentences.}
	\label{fig:anet:partial}
\end{figure}

In this section, we investigate the correlation between the number of observed clips and sentences and models' performance of video and paragraph retrieval. In this experiment, we gradually increase the number of clips and sentences observed by our model during the testing and obtained the Figure~\ref{fig:anet:partial}, on ActivityNet. When the video/paragraph contains fewer clips/sentences than the number of observations we required, we take all those available clips/sentences for computing the video/paragraph embedding. (On average 3.65 clips/sentences per video/paragraph)

From Figure~\ref{fig:anet:partial}, we note that increasing the number of the observed clips and sentences leads to improved performance results in retrievals. We can see that when observing only one clip and sentence, our model already outperforms the previous state-of-the-art method as well as our baseline \fse that observes the entire sequence. With observing less than the average length of clips and sentences, our learned model can achieve $\sim 70\%$ of the final performance.

\subsection{Results on Video Captioning}
\label{exp:caption}

\subsubsection{Setups.} In addition to the video paragraph retrieval, we evaluate our learned embeddings for video captioning. Specifically, we follow ~\cite{krishna2017dense} and train a caption model~\cite{vinyals2015show} on top of the pre-trained video embeddings.  Similar to~\cite{krishna2017dense}, we concatenate the clip-level feature with contextual video-level feature, and build a two-layer LSTM as a caption generator. We randomly initialized the word embedding as well as LSTM and trained the model for 25 epochs with learning rate of 0.001. We use the ground-truth proposal throughout training and evaluation following the setting of ~\cite{krishna2017dense,li2018jointly}. During testing, beam search is used with beam=5. Results are reported in Table~\ref{tab:anet:caption}.

\begin{table}[t]
	\centering
    \begin{minipage}{.5\linewidth}
    \centering
	\small
	\caption{Results for video captioning on ActivityNet }
	\begin{tabular}{ccccccc}
		       & B@1 & B@2 & B@3 & B@4 & M & C \\ \hline
		\alg{lstm-yt}\cite{venugopalan2015sequence}     & 18.2 & 7.4 & 3.2 & 1.2 & 6.6 & 14.9 \\
		\alg{s2vt}\cite{venugopalan2014translating}     & 20.4 & 9.0 & 4.6 & 2.6 & 7.9 & 21.0 \\
		\alg{hrnn}\cite{yu2016video}                    & 19.5 & 8.8 & 4.3 & 2.5 & 8.0 & 20.2 \\
		\alg{dense}\cite{krishna2017dense}              & 26.5 & 13.5 & 7.1 & 4.0 & 9.5 & 24.6 \\
		\alg{dvc} \cite{li2018jointly}                  & 19.6 & 9.9 & 4.6 & 1.6 & 10.3 & 25.2 \\ \hline
		\fse				                            & 17.9 & 8.2 & 3.6 & 1.7 & 8.7 & 32.1 \\
		\thse                                           & 19.6 & 9.4 & 4.2 & 2.0 & 9.2 & 39.5 \\
		\hse{\scriptsize[$\tau$=5e-4]}                  & 19.8 & 9.4 & 4.3 & 2.1 & 9.2 & 39.8 \\
	\end{tabular}
	\label{tab:anet:caption}
	\end{minipage}%
	\begin{minipage}{.05\linewidth}
	    \
	\end{minipage}%
    \begin{minipage}{.45\linewidth}
	\small
	\caption{Results for action recognition on ActivityNet (low-level embeddings)}
	\begin{tabular}{c@{\quad}cccc}
		& \multicolumn{2}{c}{\textbf{Zero-Shot}} & \multicolumn{2}{c}{\textbf{Train}} \\
		& \multicolumn{2}{c}{\textbf{Transfer}} & \multicolumn{2}{c}{\textbf{Classifier}} \\
		& Top-1 & Top-5  & Top-1 & Top-5 \\ \hline
		\alg{fv-vae}~\cite{qiu2017deep} & - & - & 78.6 & - \\
		\alg{tsn}~\cite{wang2017temporal} & - & - & 88.1 & - \\ \hline
		\fse            & 48.3 & 79.4 & 74.4 & 94.1 \\
		\thse           & 50.2 & 84.4 & 74.7 & 94.3 \\
		\hse{\scriptsize[$\tau$=5e-4]} & 51.4 & 83.8 & 75.3 & 94.3 \\ \hline
		\alg{random}    & 0.5 & 2.5 & 0.5 & 2.5 \\
	\end{tabular}
	\label{tab:anet:zero-shot}
	\end{minipage}
\end{table}

\subsubsection{Results.} We observe that our proposed model outperforms baseline over most metrics. Meanwhile, \hse also improves over previous approaches such as \alg{lstm-yt}, \alg{s2vt}, and \alg{hrnn} on B@2, METEOR, and CIDEr by a margin. \hse achieves comparable results with \alg{dvc} in all criterions. However, both \hse and \thse failed to obtain close performance to \alg{dense}~\cite{krishna2017dense}. This may due to the fact that \alg{dense}~\cite{krishna2017dense} carefully learns to aggregate the context information of a video clip for producing high-quality caption, while optimized for video-paragraph retrieval our embedding model does not equip with such capability. However, it is worth noting that our model obtains higher CIDEr score compared to all existing methods. We empirically observe that fine-tuning the pre-trained video embedding does not lead to further performance improvement.

\subsection{Results on Action Recognition}
\label{exp:action_recognition}

To evaluate the effectiveness of our model, we take the off-the-shelf clip-level embeddings trained on video-paragraph retrieval for action recognition (on ActivityNet with non-overlapping training and validation data). We use two action recognition settings to evaluate, namely \textbf{zero-shot transfer} and \textbf{classification}.

\subsubsection{Setups.} In the \textbf{zero-shot} setting, we directly evaluate our low-level embedding model learned in the video and text retrieval, via treating the phrases of actions as sentences and use the sentence-level encoder to encode the action embedding. We take the raw video and apply clip-level video encoder to extract the feature for retrieving actions. No re-training is performed and all models have no access to the actions' data distribution. Note though action are not directly used as sentences during the training, some are available as verbs in the vocabulary. Meanwhile, as we are using pre-trained word vector (GloVe), it allows the transfer to unseen actions. In the \textbf{classification} setting, we discriminatively train a simple classifier to measure the classification accuracy. Concretely, a one-hidden-layer Multi-Layer Perceptron (MLP) is trained on the clip-level embeddings. We do not fine-tune the pre-trained clip-level video embedding here.

\subsubsection{Results.} We report results of above two settings on the ActivityNet validation set (see Table~\ref{tab:anet:zero-shot}). We observe that our learned low-level embeddings allow superior zero-shot transfer to action recognition, without accessing any training data. This indicates that semantics of actions are indeed well reserved in the learned embedding models. More interestingly, we can see that both \thse and \hse improve the performance over \fse. It shows that our hierarchical modeling of video benefits not only high-level embedding but also low-level embedding. A similar trend is also observed in the classification setting. Our method achieves comparable performance to the state-of-the-art video modeling approach such as \alg{fv-vae}~\cite{qiu2017deep}. Note \alg{tsn}~\cite{wang2017temporal} is fully supervised thus not directly comparable.

\subsection{Qualitative Results}

\begin{figure}[t]
	\centering
	\begin{tabular}{cc}
		\includegraphics[width=0.475\textwidth,trim={1.5cm 1.5cm 1.5cm 1.5cm},clip]{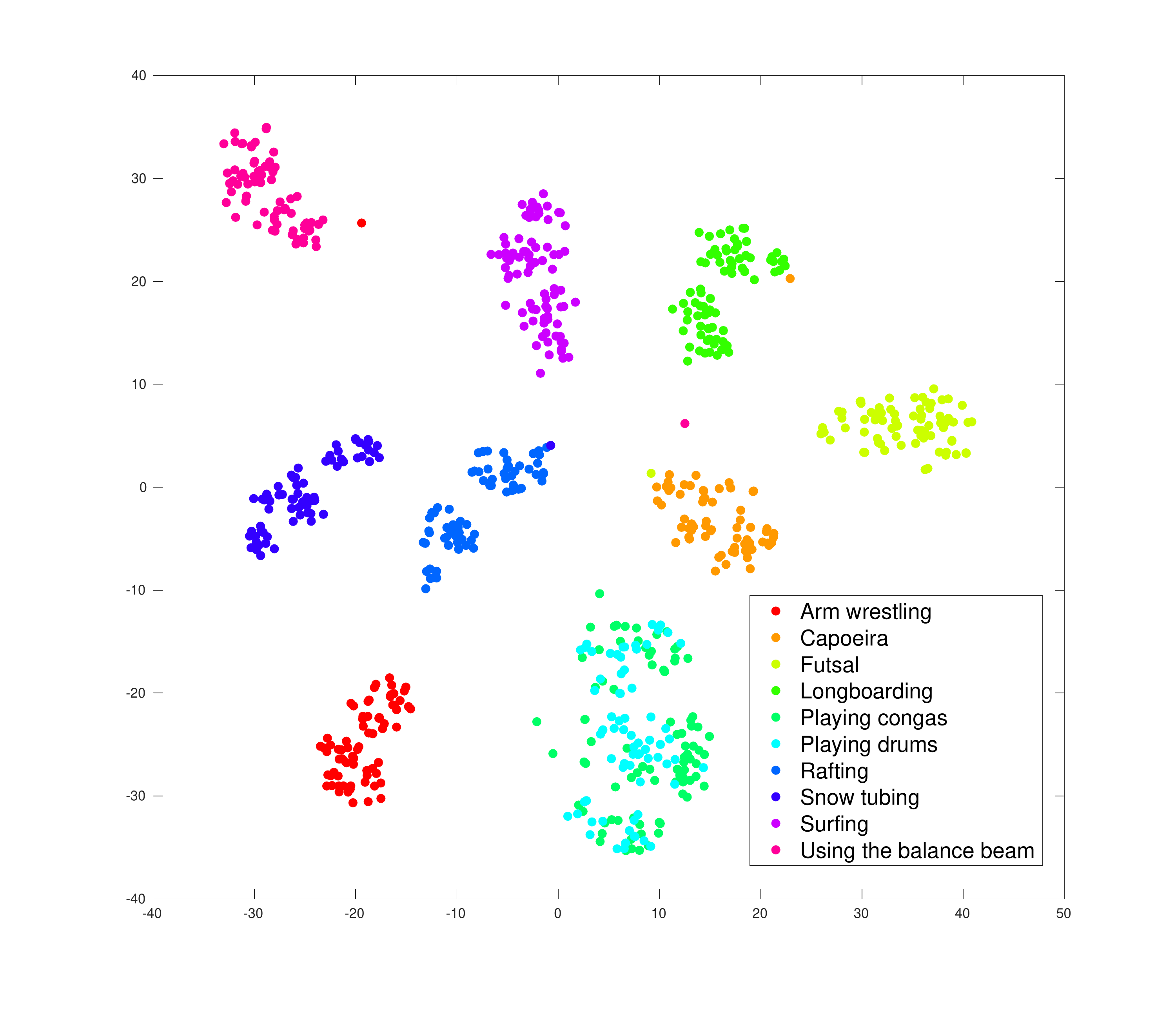} &
		\includegraphics[width=0.475\textwidth,trim={1.5cm 1.5cm 1.5cm 1.5cm},clip]{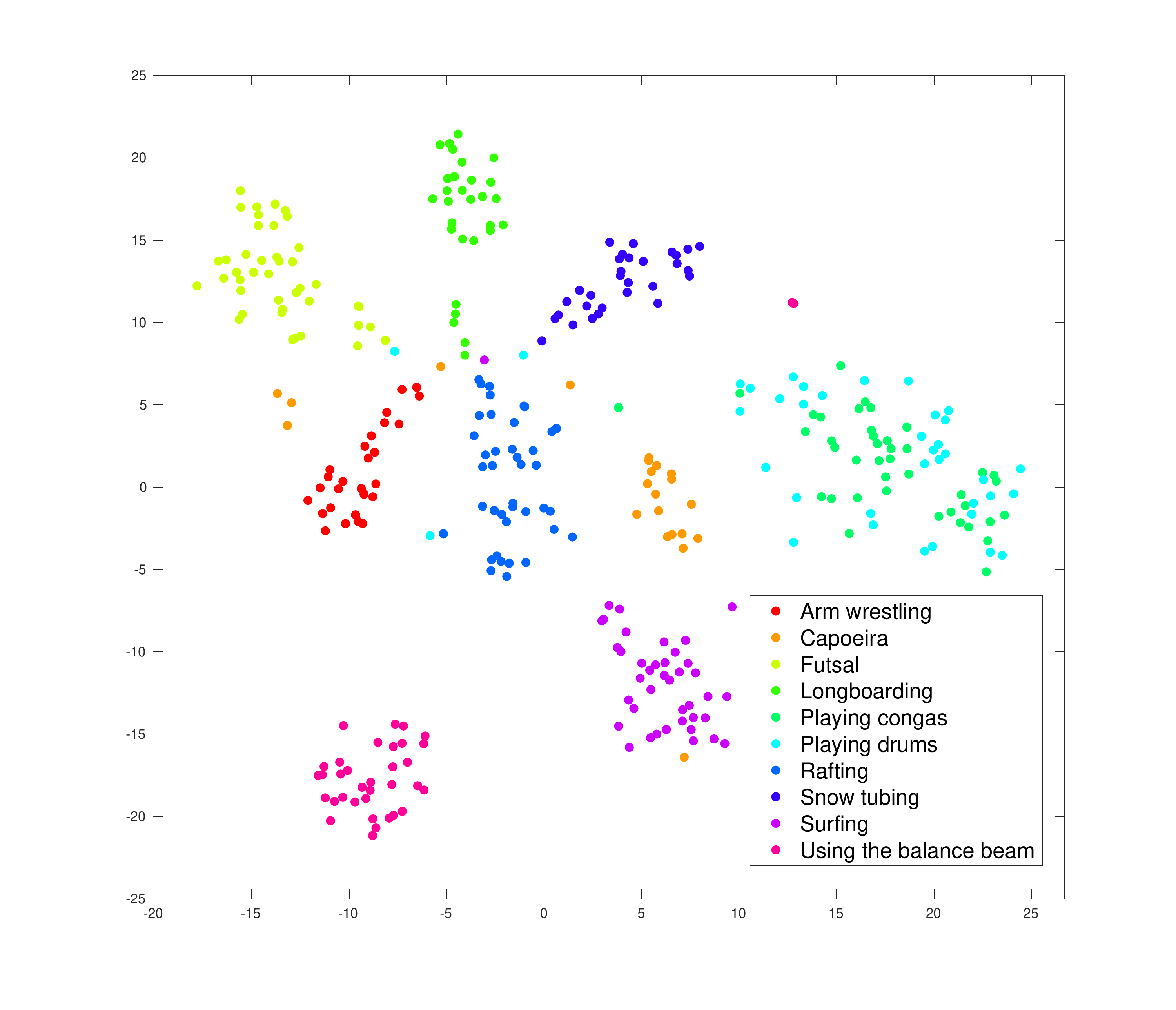} \\
		ActivityNet Training Set & ActivityNet Validation Set
	\end{tabular}
	\caption{T-SNE visualization of off-the-shelf video embedding of \hse on ActivityNet v1.3 training and validation set. Points are marked with its action classes.}
	\label{fig:tsne}
\end{figure}

We use t-SNE~\cite{maaten2008visualizing} to visualize our results in the video to paragraph and paragraph to video retrieval task. Fig~\ref{fig:tsne} shows that the proposed method can cluster the embedding of videos with regard to its action classes.
To further explain the retrieval quality, we provide qualitative visualization in the Supp. Material.

% !TEX root = Arxiv_ECCV_BZ_HH_FS.tex
\section{Conclusion}
\label{conclusion}

In this paper, we propose a novel cross-modal learning approach to model videos and texts jointly, which leverages the intrinsic hierarchical structures of both videos or texts. Specifically, we consider the correspondences of videos and texts at multiple granularities, and derived loss functions to align the embeddings for the paired clips and sentences, as well as paired video and paragraph in accordance in their own semantic spaces. Another important component of our model is layer-wise reconstruction, which ensures that learned embeddings capture video (paragraph) and clips (words) at different levels. Moreover, we further extend our learning objective so that it allows to handle a more generalized learning scenario where only video paragraph correspondence exists. We demonstrate the advantage of our proposed model in a range of tasks including video and text retrieval, zero-shot action recognition and video caption.

\hfill\break
\noindent
\textbf{Acknowledgments} {\small We appreciate the feedback from the reviewers. This work is partially supported by NSF IIS-1065243, 1451412, 1513966/ 1632803/1833137, 1208500, CCF-1139148, a Google Research Award, an Alfred P. Sloan Research Fellowship, gifts from Facebook and Netflix, and ARO\# W911NF-12-1-0241 and W911NF-15-1-0484.}

\bibliographystyle{splncs04}
\bibliography{refs_ECCV}
\newpage
\appendix

\section{Implementation Details}
\label{sec:sup:impl}

\subsection{Video and Text Features}

\subsubsection{C3D Features.}
Similar to ~\cite{krishna2017dense}, we follow the standard ActivityNet setting and use the C3D~\cite{tran2015learning} features from~\cite{heilbron2015activitynet} for retrieval and dense captioning~\cite{krishna2017dense}. In all our experiments under this setting, we extract frame-wise video feature using C3D model pre-trained on Sports-1M dataset, with the temporal stride of 16. PCA dimensionality reduction is then conducted to reduce features dimension to 500.

\subsubsection{TSN-Inception V3 Features.}
To leverage the state-of-the-art of current video modeling, we extract more recent deep features for retrieval on ActivityNet~\cite{krishna2017dense} and DiDeMo~\cite{anne2017localizing}, using the Inception V3 model pre-trained on Kinetics~\cite{kay2017kinetics} dataset (provided by ~\cite{wang2016temporal}).  Follow their settings, we resize video frames to the resolution of $299 \times 299$. We then fed video frames into the deep Inception V3 model to extract the output activations from penultimate layer. Unlike~\cite{wang2016temporal}, we do not perform any test-time data augmentations (\eg multiple crops, color jitter, etc.). Note that no fine-tuning are performed on either ActivityNet or DiDeMo.

\subsubsection{Word Features.}
In the retrieval related experiments, we always use GloVE features~\cite{pennington2014glove} for the initialization of the word embedding and fine-tune. Specifically, we use the GloVE vectors pre-trained on 840B common web-crawled data, with its dimensionality equals to 300.

\subsubsection{Training Details}
When the learning of hierarchical embedding is applicable, we feed the entire video/paragraph in its frame-wise/word-wise representations through the low-level encoder, and then input the subsequent low-level embedding to the high-level encoder as its initial hidden state. In all our experiments, we use GRU~\cite{cho2014learning} with its hidden dimension to be 1,024 as our sequence encoder and decoder. To obtain the embedding for a sequence, we take the channel-wise max over all output vectors of the GRU as it empirically outperforms other strategies such as ~\cite{venugopalan2015sequence}.

During training, we use Adam~\cite{kingma2014adam} optimizer with initial learning rate as 0.001, and decay it by 10 for every 10 epochs during the training. We use Xavier initialization~\cite{glorot2010understanding} for each affine layer in our model with zero mean and variance of 0.01. We set all margin in the loss function to $0.2$. Each loss is normalized by its batch size. On both ActivityNet and DiDeMo dataset, we train our embedding models for 15 epochs and collect the final results.

\section{Additional Experiments}
\label{sec:sup:exp}
\subsection{Ablation Study on Different Learning Objectives}

\subsubsection{Ablation Study with Different Learning Objectives} We report ablation studies of different losses on ActivityNet video and paragraph retrieval task in Table \ref{tab:ablation_supp}. We use the Inception-V3 features and follow the same setting for training \alg{hse}. Each time we remove one loss and report the performance. Note that the reconstruction loss and low-match loss are the most useful.

\begin{table}[ht]
	\centering
	\small
	\setlength{\tabcolsep}{10pt}
	\caption{Ablation study on the learning objectives. }
	\begin{tabular}{lcccc}
		& \multicolumn{2}{c}{\text{ \textbf{P}aragraph } $\Rightarrow$ \text{ \textbf{V}ideo }} & \multicolumn{2}{c}{\text{ \textbf{V}ideo } $\Rightarrow$ \text{ \textbf{P}aragraph}} \\
		Method & R@1 & R@5 & R@1 & R@5 \\ \hline
		\hse w/o high-cluster  &  44.6 & 76.4 & 44.2 &  76.1   \\
		\hse w/o low-match & 40.9 & 73.6 & 39.8 & 73.6   \\
		\hse w/o low-cluster  & 44.6 & 76.6 & 43.9 & 76.4 \\
		\hse w/o reconstruction & 43.9 & 75.8 & 43.3 & 75.3 \\
		\hse w all losses & 44.4 & 76.7 & 44.2 & 76.7\\
	\end{tabular}
	\label{tab:ablation_supp}
\end{table}

\subsubsection{Low-level Loss is Beneficial} 
\begin{figure}[ht]
	\small
	\centering
	\begin{tabular}{cccc}
		\includegraphics[width=0.225\textwidth]{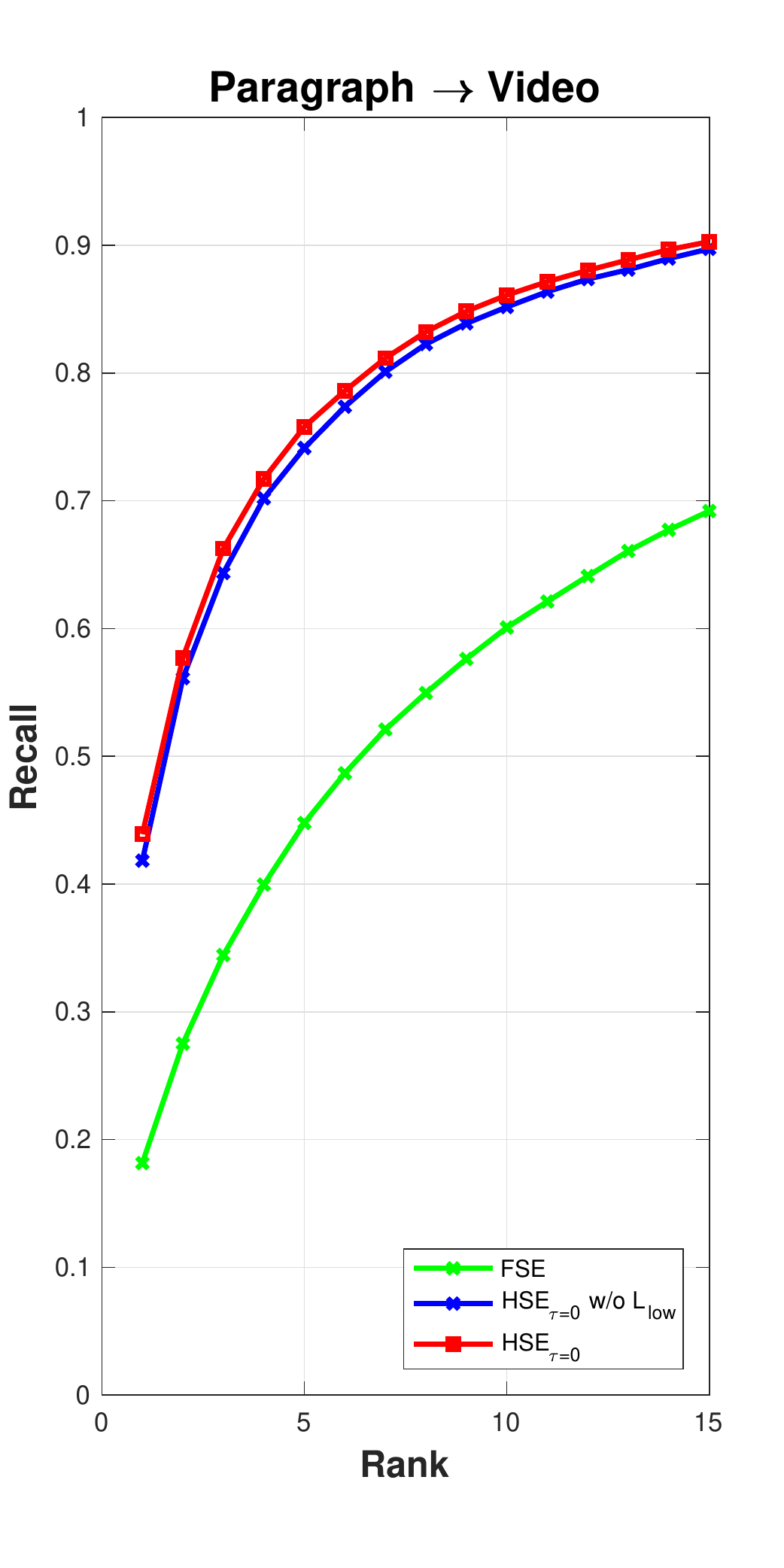} &
		\includegraphics[width=0.225\textwidth]{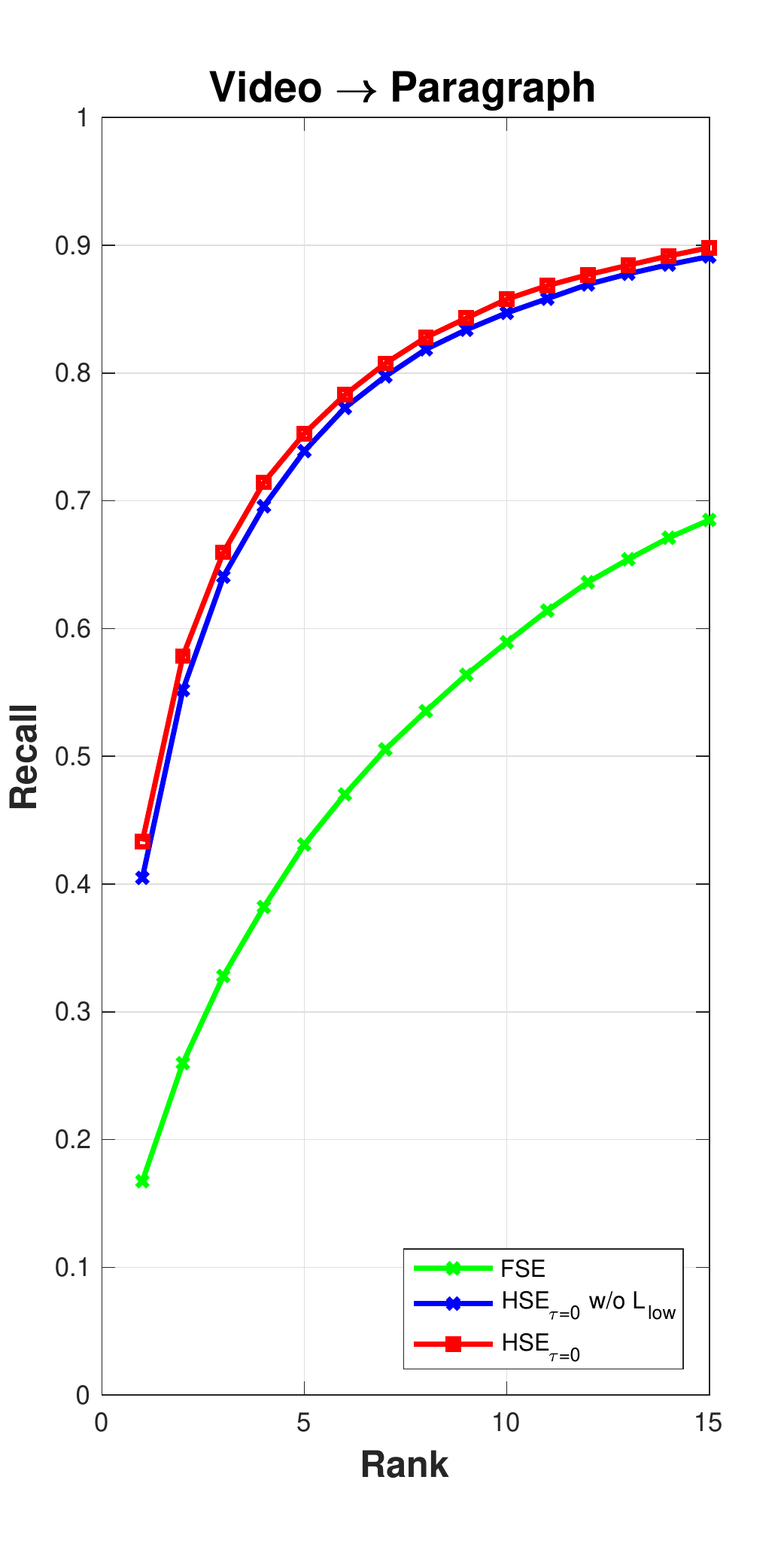} &
		\includegraphics[width=0.225\textwidth]{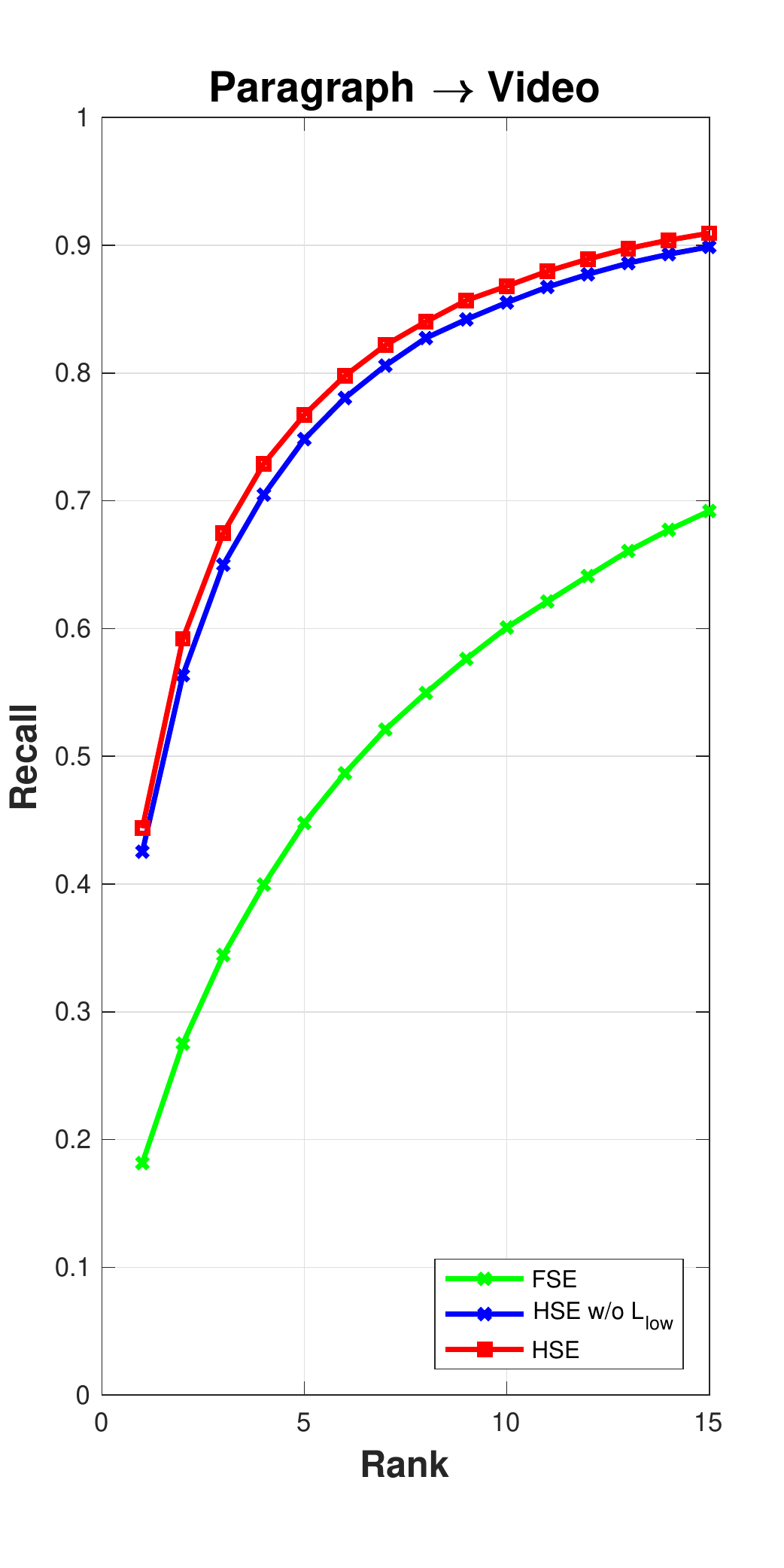} &
		\includegraphics[width=0.225\textwidth]{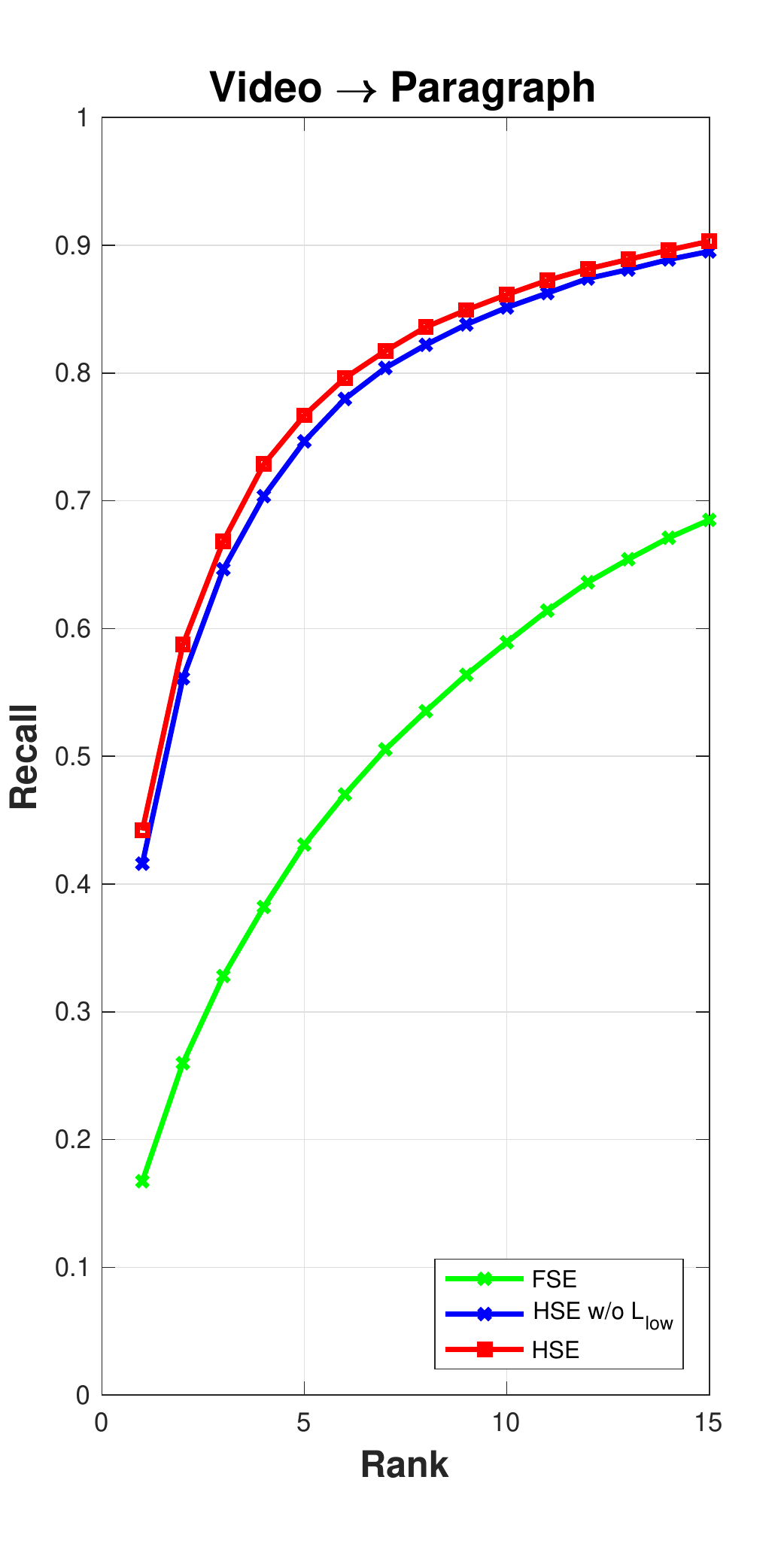} \\
		\multicolumn{2}{c}{(a) \thse} &
		\multicolumn{2}{c}{(b) \hse}
	\end{tabular}
	\caption{Recall vs Rank curves of Video to Paragraph and Paragraph to Video retrieval of both \thse and \hse. All results are collected from models based on InceptionV3 feature on ActivityNet validation set 1.}
	\label{fig:RRcurve}
\end{figure}

As mentioned in the main text (see Table 1 and Table 2 in the main text), learning with low-level objectives is beneficial for our full model. To better understands this, we also plot the recall (in $\%$) with regard to the rank of the video/paragraph to a query as supportive evidence. The results are shown in Fig. ~\ref{fig:RRcurve}.

\subsection{Ablation Study on Reconstruction Balance Term}
Here we study the influence of loss balance term, by experimenting multiple choices of $\tau$ under a controlled environment. We choose to study this on the validation set 2 (val2) of ActivityNet with Inception V3 visual feature as input. Detailed results are shown in Table~\ref{tab:tau}. We summarized that retrieval performance, R@1 and R@5, approach to its peak when $\tau$=0.0005. Therefore, as stated in the main text, we set $\tau$ to be 0.0005 in all our experiments.
\begin{table}[ht]
	\centering
	\small
	\setlength{\tabcolsep}{3pt}
	\caption{Ablation study of $\tau$ on ActivityNet (val2).}
	\begin{tabular}{ccccccccc}
		& \multicolumn{4}{c}{\text{ \textbf{P}aragraph } $\Rightarrow$ \text{ \textbf{V}ideo }} & \multicolumn{4}{c}{\text{ \textbf{V}ideo } $\Rightarrow$ \text{ \textbf{P}aragraph}} \\
		& R@1 & R@5 & R@50 & MR & R@1 & R@5 & R@50 & MR \\ \hline
		\multicolumn{9}{c}{Inception-V3 pre-trained on Kinetics~\cite{wang2017temporal}} \\ \hline
		\textsc{hse}\scriptsize{[$\tau$=0.05]}& 25.0 & 54.9 & 92.6 & 5.0 & 25.1 & 55.4 & 92.4 & 4.0   \\
		\textsc{hse}\scriptsize{[$\tau$=0.005]}  & 32.4 & 62.2 & 93.8 & 3.0 & 32.1 & 63.0 & 93.7 & 3.0\\
		\textsc{hse}\scriptsize{[$\tau$=0.0005]} & \textbf{33.2} & \textbf{62.9} & \textbf{93.6} & \textbf{3.0} & \textbf{32.6} & \textbf{62.8} & \textbf{93.5} & \textbf{3.0}  \\
		\textsc{hse}\scriptsize{[$\tau$=0.00005]} & 33.2 & 62.9 & 93.8 & 3.0 & 32.2 & 62.5 & 93.6 & 3.0  \\
		\textsc{hse}\scriptsize{[$\tau$=0]} & 32.2 & 61.5 &	93.6 & 3.0 & 31.5 &	62.0 & 93.3 & 3.0 \\
	\end{tabular}
	\label{tab:tau}
\end{table}

\subsection{Performance on ActivityNet Validation Set 2}
As mentioned in the main paper, we reported the val2 performance of \fse, \thse, and \hse in Table~\ref{tab:anet_val2}. Again, the results verified our papers' claim as we show that \hse consistently improve performance than \fse and \thse. It shows the importance of hierarchical modeling and feature reconstruction.

\begin{table}[ht]
	\centering
	\small
	\setlength{\tabcolsep}{3pt}
	\caption{Performance of video and paragraph retrieval on ActivityNet (val2). Standard deviation from 3 random seeded experiments are also reported.}
	\begin{tabular}{ccccccccc}
		& \multicolumn{4}{c}{\text{ \textbf{P}aragraph } $\Rightarrow$ \text{ \textbf{V}ideo }} & \multicolumn{4}{c}{\text{ \textbf{V}ideo } $\Rightarrow$ \text{ \textbf{P}aragraph}} \\
		& R@1 & R@5 & R@50 & MR & R@1 & R@5 & R@50 & MR \\ \hline
		\multicolumn{9}{c}{C3D Feature with Dimensionality Reduction~\cite{tran2015learning}} \\ \hline
		\fse  & 11.5\std{0.2} & 31.0\std{0.4} &	75.9\std{0.2} & 14.0 & 11.0\std{0.5} & 30.6\std{0.3}	& 75.5\std{0.4}	& 14.0  \\
		\thse & 23.3\std{0.5} & 48.2\std{0.2} & 84.5\std{0.4} & 6.0  & 23.0\std{0.3} & 47.9\std{0.2} & 84.6\std{0.2} & 6.0 \\
		\hse{\scriptsize[$\tau$=0.0005]} & 23.9\std{0.3} & 49.4\std{0.3} & 85.3\std{0.2} & 6.0 & 23.4\std{0.5} & 49.4\std{0.4} & 85.5\std{0.3} & 6.0 \\ \hline

		\multicolumn{9}{c}{Inception-V3 pre-trained on Kinetics~\cite{wang2017temporal}} \\ \hline
		\fse & 16.0\std{0.2} &  41.8\std{0.4} & 88.0\std{0.5} & 8.0 & 15.1\std{0.7}  & 41.0\std{0.4}  & 87.7\std{0.5}  & 8.0  \\
		\thse & 32.3\std{0.2} & 62.2\std{0.7} & 93.5\std{0.1} & 3.0 & 32.0\std{1.0} & 61.9\std{0.2} & 93.3\std{0.1} & 3.0 \\
		\hse{\scriptsize[$\tau$=0.0005]} & \textbf{32.9\std{0.4}} & \textbf{62.7\std{0.2}} & \textbf{93.9\std{0.4}} & \textbf{3.0} & \textbf{32.6\std{0.1}} & \textbf{63.0\std{0.2}} & \textbf{93.7\std{0.2}} & \textbf{3.0} \\
	\end{tabular}
	\label{tab:anet_val2}
\end{table}

\section{Visualization}
\label{sec:sup:vis}
\begin{figure}[t]
	\centering
	\begin{tabular}{c|l}
		\alg{query video} & \raisebox{-.5\height}{\includegraphics[width=0.785\textwidth]{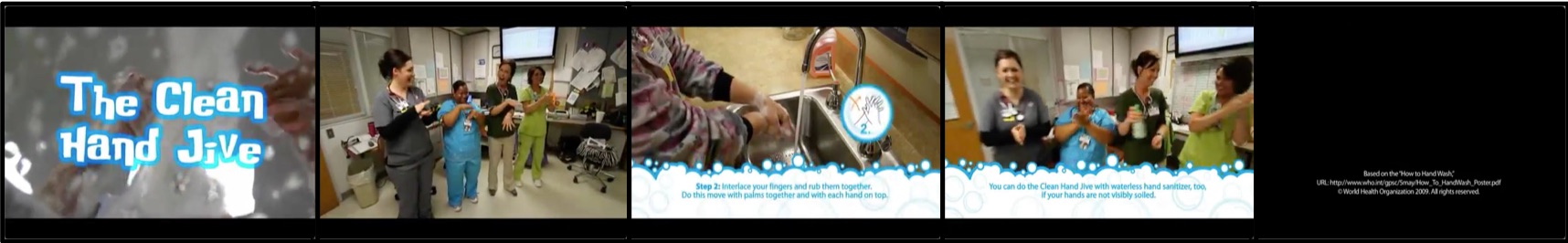}} \\
	\alg{ground truth} &	\cfbox{olive}{
													\begin{tabular}{c}
														\multicolumn{1}{m{0.755\textwidth}}{
															\tiny \textbf {
																\color{black} {
																	The credits of the clip are shown. People clean their hand and use their hands to dance. The credits of the video are shown.
																}
															}
														}
													\end{tabular}
													} \\
	[2ex]
	\hse &
	\cfbox{olive}{
		\begin{tabular}{c}
				\multicolumn{1}{m{0.755\textwidth}}{
					\tiny \textbf {
						\color{black} {
							The credits of the clip are shown. People clean their hand and use their hands to dance. The credits of the video are shown.
						}
					}
				}
			\end{tabular}
			} \\
	[2ex]
	\thse &
	\cfbox{olive}{
		\begin{tabular}{c}
			\multicolumn{1}{m{0.755\textwidth}}{
				\tiny \textbf {
					\color{black} {
					The credits of the clip are shown. People clean their hand and use their hands to dance. The credits of the video are shown.
					}
				}
			}
		\end{tabular}
	} \\
	[2ex]
	\fse &
	\cfbox{red}{
		\begin{tabular}{c}
			\multicolumn{1}{m{0.755\textwidth}}{\tiny \textbf {\color{black} { Two women are seen speaking to the camera and leads into turning on a faucet and running her hands underneath. The woman then scrubs soap into her hands and continues to wash them off then taking a paper down and drying her hands and sink. The other steps in to demonstrate how she washes her hands and ends by laughing to the camera.}}}
		\end{tabular}
		} \\
	\hline
	\end{tabular}
	\begin{tabular}{c|l}
		\alg{query video} &\raisebox{-.5\height}{\includegraphics[width=0.785\textwidth]{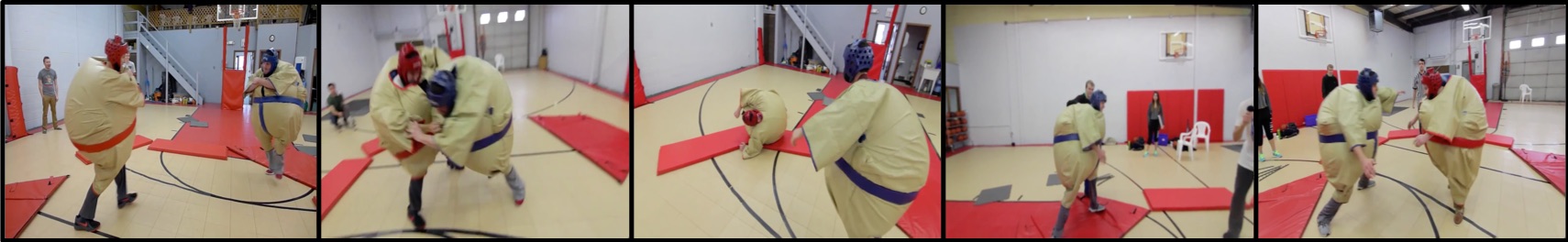}} \\
		\alg{ground truth} &
	\cfbox{olive}{
		\begin{tabular}{c}
			\multicolumn{1}{m{0.755\textwidth}}{\tiny \textbf {\color{black} { Two people dressed up in sumo wrestler suits come running into a gym and wrestle while people stand around and watch. The wrestler wearing red falls over. They continue wrestling and are having a lot of fun doing it falling down and bouncing around. One of the wrestlers wearing blue makes a shot in a basketball net. The two people continue wrestling in their sumo suits. A man comes into the shot and pushes the sumo wrestler over on top of another person not wearing a suit. There comes a final of the sumo wrestlers and a man in a white shirt is presenting and there is a referee. They start on the wrestle while people watch swinging each other around in the middle of the red mats. The red sumo wrestler falls down and the blue sumo wrestler wins. The blue sumo wrestler jumps up happy with his friends and walks out the door and the red sumo wrestler is left on the ground.
}}}
		\end{tabular}
	} \\
	[9ex]
	\hse  &
	\cfbox{olive}{
		\begin{tabular}{c}
			\multicolumn{1}{m{0.755\textwidth}}{\tiny \textbf {\color{black} { Two people dressed up in sumo wrestler suits come running into a gym and wrestle while people stand around and watch. The wrestler wearing red falls over. They continue wrestling and are having a lot of fun doing it falling down and bouncing around. One of the wrestlers wearing blue makes a shot in a basketball net. The two people continue wrestling in their sumo suits. A man comes into the shot and pushes the sumo wrestler over on top of another person not wearing a suit. There comes a final of the sumo wrestlers and a man in a white shirt is presenting and there is a referee. They start on the wrestle while people watch swinging each other around in the middle of the red mats. The red sumo wrestler falls down and the blue sumo wrestler wins. The blue sumo wrestler jumps up happy with his friends and walks out the door and the red sumo wrestler is left on the ground.
}}}
		\end{tabular}
	} \\
	[9ex]
	\thse &
	\cfbox{red}{
		\begin{tabular}{c}
			\multicolumn{1}{m{0.755\textwidth}}{\tiny \textbf {\color{black} { There are some girls wearing karate uniforms doing karate on a stage. There's an orange belt and a yellow belt karate student doing some karate moves with batons in their hands. After they leave, another karate student wearing a yellow belt comes on stage to perform her karate moves with a baton. Then she leaves and another girl wearing an orange belt joins in holding two hammers to show her karate moves. She leaves and another girl wearing a yellow belt comes on stage with a hand fan and shows her karate moves. After she leaves the master comes on stage along with three other students. They take turns to smash the board held by the master. Then the master leaves and the three students demonstrate their coordinated karate moves.
}}}
		\end{tabular}
	} \\
	[7.5ex]
	\fse &
	\cfbox{red}{
		\begin{tabular}{c}
			\multicolumn{1}{m{0.755\textwidth}}{\tiny \textbf {\color{black} { A girl walks along the gym holding her fencing gear. She points at the camera with her sword. A coach comes to dress her and fix equipment. The gym is full of kids fencing and practicing. She starts fencing with another girl. A coach in a blue shirt gives the direction.
}}}
		\end{tabular}
		} \\
	\end{tabular}

	\caption{{\small \textbf{ActivityNet: Given Video and Retrieve Paragraph.}
  Positive qualitative examples of \hse, \thse, and \fse on the task of given video to retrieve texts. We mark the correct sample in \textbf{ \color{olive} green} and incorrect one in \textbf{ \color{red} red}. } }
	\label{anet_v2p_pos}
\end{figure}

\begin{figure}[t]
	\centering
	\begin{tabular}{c|l}
		\alg{query video} &
		\raisebox{-.5\height}{\includegraphics[width=0.785\textwidth]{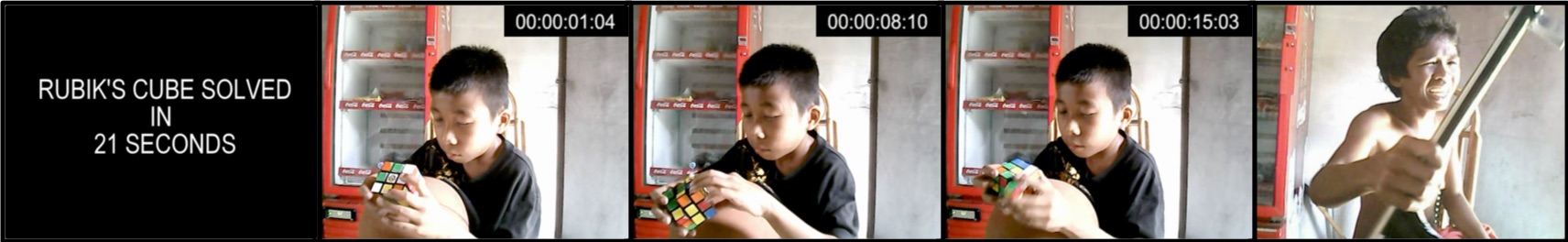}} \\
	    \alg{ground truth} &
	    \cfbox{olive}{\begin{tabular}{c}
			\multicolumn{1}{m{0.772\textwidth}}{\tiny \textbf {\color{black} {A child solves a cube puzzle while holding a basketball. A hand holding an object touch the ear of the boy. The boy stand and leave, then a naked man appears holding a rod.}}}
			\end{tabular}} \\
	[2.5ex]
	    \hse &
	    \cfbox{red}{\begin{tabular}{c}
			\multicolumn{1}{m{0.772\textwidth}}{\tiny \textbf {\color{black} {A woman tries to solves a cube puzzle. After, the woman pass the cube to a man that starts to solves the puzzle. The man solves the puzzle.}}}
			\end{tabular}} \\
	[2ex]
	    \thse &
	    \cfbox{olive}{\begin{tabular}{c}
			\multicolumn{1}{m{0.772\textwidth}}{\tiny \textbf {\color{black} {A child solves a cube puzzle while holding a basketball. A hand holding an object touch the ear of the boy. The boy stand and leave, then a naked man appears holding a rod.}}}
			\end{tabular}} \\
	[2.5ex]
	    \fse &
	    \cfbox{red}{ \begin{tabular}{c}
			 \multicolumn{1}{m{0.763\textwidth}}{\tiny \textbf {\color{black} {A sitting man holds a pack of cigarettes and a lighter to the camera. The man puts a cigarette in his mouth and smokes the cigarette fast. The man removes the cigarette and puts it back. The man shows the camera the end of the cigarette and smokes the rest. The man removes the cigarette and shows it to the camera.}}}
			 \end{tabular}} \\
	\hline
	\end{tabular}
	\begin{tabular}{c|l}
		\alg{query video} &\raisebox{-.5\height}{\includegraphics[width=0.785\textwidth]{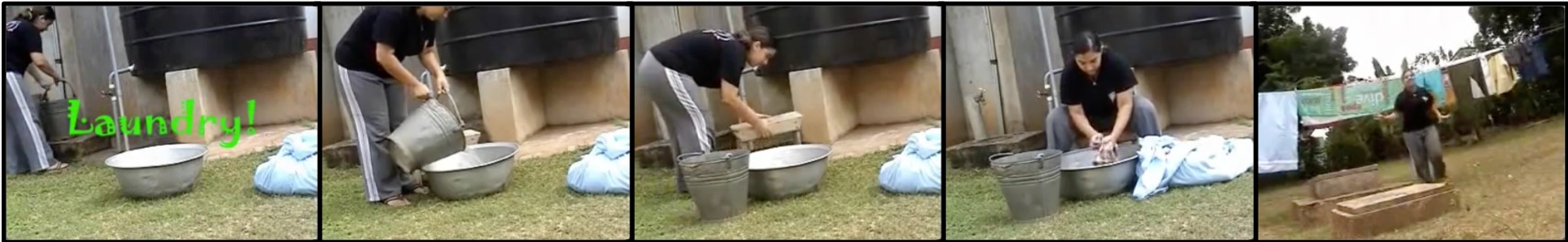}} \\
		\alg{ground truth} &
	\cfbox{olive}{ \begin{tabular}{c}
		\multicolumn{1}{m{0.763\textwidth}}{\tiny \textbf {\color{black} {A woman is seen turning on a hose to pour water in and walks over to another bucket and sits down. The woman then washes the clothes in the bucket while looking to the camera leading into her hanging up the clothes and speaking to the camera.}}}
		\end{tabular}} \\
	[3ex]
	\hse  &
	\cfbox{red}{ \begin{tabular}{c}
		\multicolumn{1}{m{0.763\textwidth}}{\tiny \textbf {\color{black} {Two kids are washing their clothes in a sink. They put the clean clothes in a bucket on the floor. }}}
		\end{tabular}} \\
	[2ex]
	\thse &
	\cfbox{red}{ \begin{tabular}{c}
		\multicolumn{1}{m{0.763\textwidth}}{\tiny \textbf {\color{black} {A man is seen kneeling over a bucket dipping clothes inside and washing. Another man is seen standing in front of a hose and dipping clothes underneath as well.}}}
		\end{tabular}} \\
	[2ex]
	\fse &
	\cfbox{red}{ \begin{tabular}{c}
		\multicolumn{1}{m{0.763\textwidth}}{\tiny \textbf {\color{black} {This man is sitting in a chair outdoors in his yard and he is washing the black shirt in the bucket. There's also 3 other buckets that have clothes in them and there are 2 other people outside. This video has no audio by the way and when the man is done, he stops to say a few words and the camera cuts off.}}}
		\end{tabular}} \\
	\end{tabular}

	\caption{{\small \textbf{ActivityNet: Given Video and Retrieve Paragraph.} Negative qualitative examples of \hse, \thse, and \fse on the task of given video to retrieve texts. We mark the correct sample in \textbf{ \color{olive} green} and incorrect one in \textbf{ \color{red} red}. } }
	\label{anet_v2p_neg}
\end{figure}

\begin{figure}[t]
	\centering
	\begin{tabular}{c|l}

	\alg{query text} & \multicolumn{1}{m{0.785\textwidth}}{\tiny A man is floating in the water holding a table and a stool. The man stands on the table and sits the stool upright. The man sits on the stool as he water skis on the table. The man stands on top of the stool then stands up. The man is standing on a stool as he water skis in  lake. The man does a spin while on the stool. The man jumps in the water as the boat drives on.} \\
		\alg{ground truth} &
		\cfbox{olive}{\raisebox{-.5\height}{\includegraphics[width=0.785\textwidth]{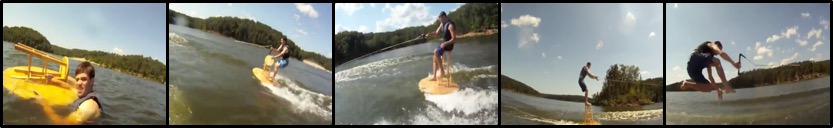}}} \\
		[5.5ex]
	\hse  &
		\cfbox{olive}{\raisebox{-.5\height}{\includegraphics[width=0.785\textwidth]{figures_supp/Anet_p2v_3/1.jpg}}} \\
		[5.5ex]
	\thse &
		\cfbox{olive}{\raisebox{-.5\height}{\includegraphics[width=0.785\textwidth]{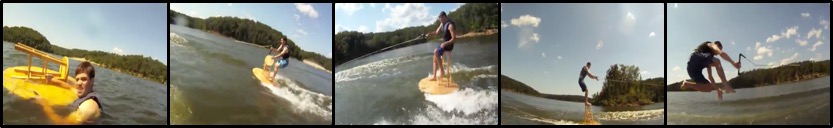}}} \\
		[5.5ex]
	\fse &
		\cfbox{red}{\raisebox{-.5\height}{\includegraphics[width=0.785\textwidth]{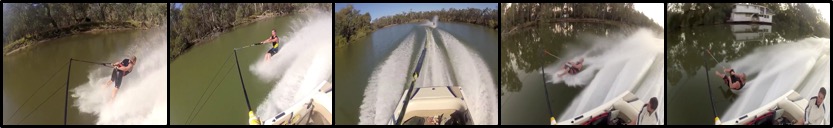}}} \\
		[5.5ex]
		\hline
	\alg{query text} & \multicolumn{1}{m{0.785\textwidth}}{\tiny We see the scoreboard of a racing game. The race starts ant the player is playing on jet skis. The player passes the red bridge. The player passes the cruise ship and zeppelin. The player passes the cliff with the lighthouse. The timer counts down from 10 and the race is finishes. We see the players record score. We see the ranking screen for the game, the level, and the option to change the difficulty.} \\
		\alg{ground truth} &
		\cfbox{olive}{\raisebox{-.5\height}{\includegraphics[width=0.785\textwidth]{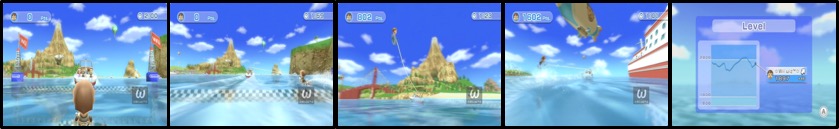}}} \\
		[5.5ex]
	\hse  &
		\cfbox{olive}{\raisebox{-.5\height}{\includegraphics[width=0.785\textwidth]{figures_supp/Anet_p2v_1/1.jpg}}} \\
		[5.5ex]
	\thse &
		\cfbox{red}{\raisebox{-.5\height}{\includegraphics[width=0.785\textwidth]{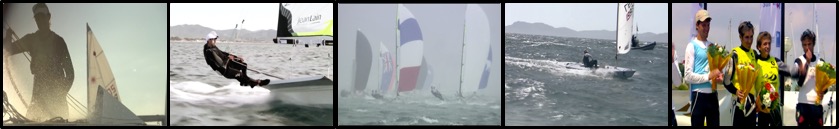}}} \\
		[5.5ex]
	\fse &
		\cfbox{red}{\raisebox{-.5\height}{\includegraphics[width=0.785\textwidth]{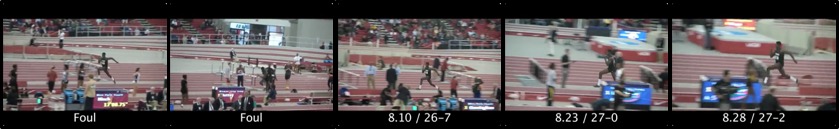}}} \\
		[5.5ex]
	\end{tabular}
	\caption{{\small \textbf{ActivityNet: Given Paragraph and Retrieve Video.} Positive qualitative examples of \hse, \thse, and \fse on the task of given text to retrieve video. We mark the correct sample in \textbf{ \color{olive} green} and incorrect one in \textbf{ \color{red} red}.} }
	\label{anet_p2v_pos}
\end{figure}

\begin{figure}[t]
	\centering
	\begin{tabular}{c|l}

	\alg{query text} & \multicolumn{1}{m{0.785\textwidth}}{\tiny A man and a woman are dancing together. The man dips under the woman's arm. The man has his hand on the woman's waist. The man puts his hand on his waist. The man hits his head accidentally. A person hits the item hanging from the roof.} \\
		\alg{ground truth} &
		\cfbox{olive}{\raisebox{-.5\height}{\includegraphics[width=0.785\textwidth]{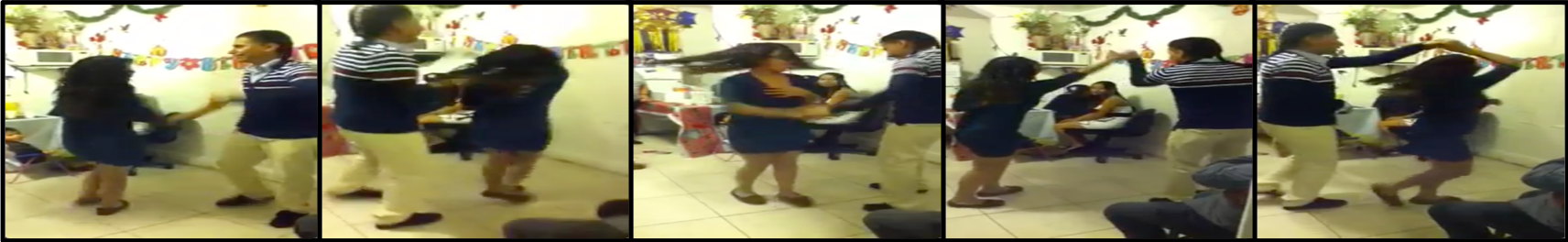}}} \\
		[5.5ex]
	\hse  &
		\cfbox{red}{\raisebox{-.5\height}{\includegraphics[width=0.785\textwidth]{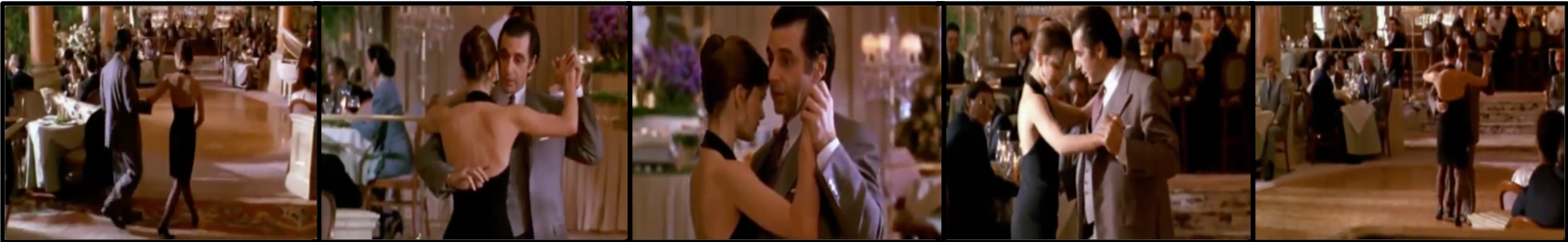}}} \\
		[5.5ex]
	\thse &
		\cfbox{olive}{\raisebox{-.5\height}{\includegraphics[width=0.785\textwidth]{figures_supp/Anet_p2v_fail_succ/1.png}}} \\
		[5.5ex]
	\fse &
		\cfbox{red}{\raisebox{-.5\height}{\includegraphics[width=0.785\textwidth]{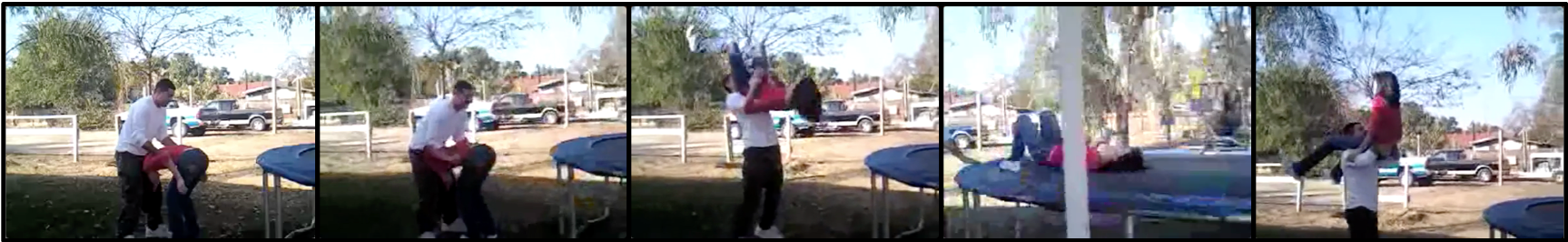}}} \\
		[5.5ex]
		\hline
	\alg{query text} & \multicolumn{1}{m{0.785\textwidth}}{\tiny A man throws a bowling ball. He then goes back and high fives his friends. They sit and talk around the table. A woman stands up and grabs a bowling ball. She walks up and drops it down the lane. She sits back down and looks at her phone. They continue talking around the table. She stands back up and picks up a bowling ball. She throws it down the lane again. She sits back down at the table. She throws a ball behind her while walking away. She picks up a ball and throws it with her hands over her eyes. She throws a bowling ball while talking on the phone.} \\
		\alg{ground truth} &
		\cfbox{olive}{\raisebox{-.5\height}{\includegraphics[width=0.785\textwidth]{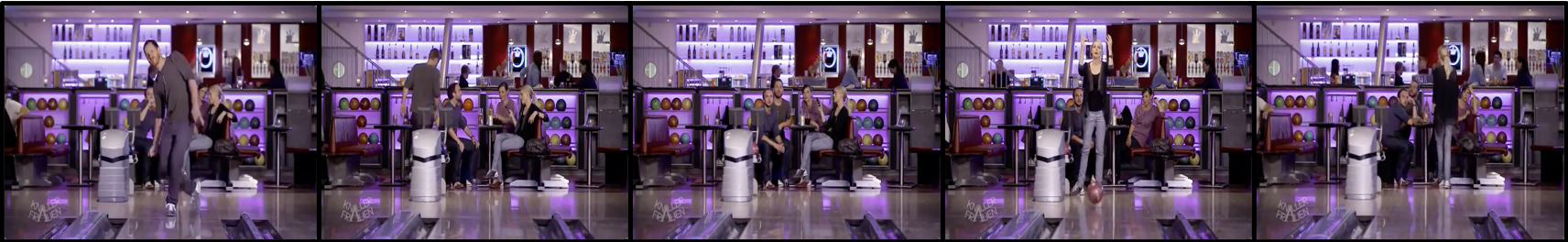}}} \\
		[5.5ex]
	\hse  &
		\cfbox{red}{\raisebox{-.5\height}{\includegraphics[width=0.785\textwidth]{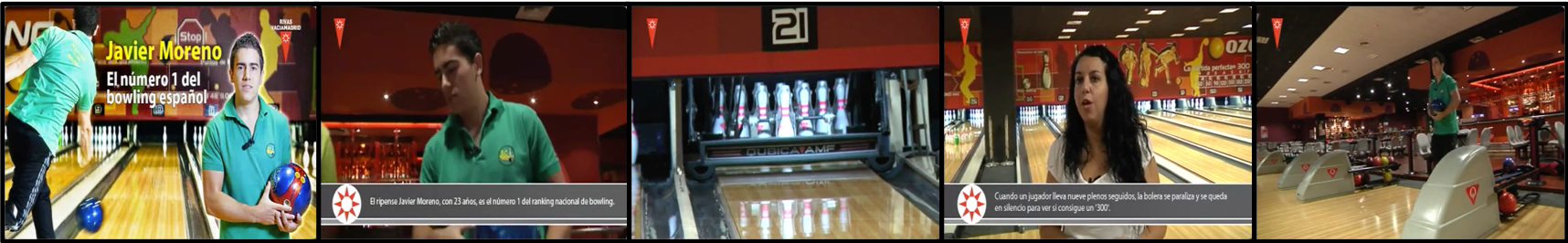}}} \\
		[5.5ex]
	\thse &
		\cfbox{red}{\raisebox{-.5\height}{\includegraphics[width=0.785\textwidth]{figures_supp/Anet_p2v_fail_fail/2.png}}} \\
		[5.5ex]
	\fse &
		\cfbox{red}{\raisebox{-.5\height}{\includegraphics[width=0.785\textwidth]{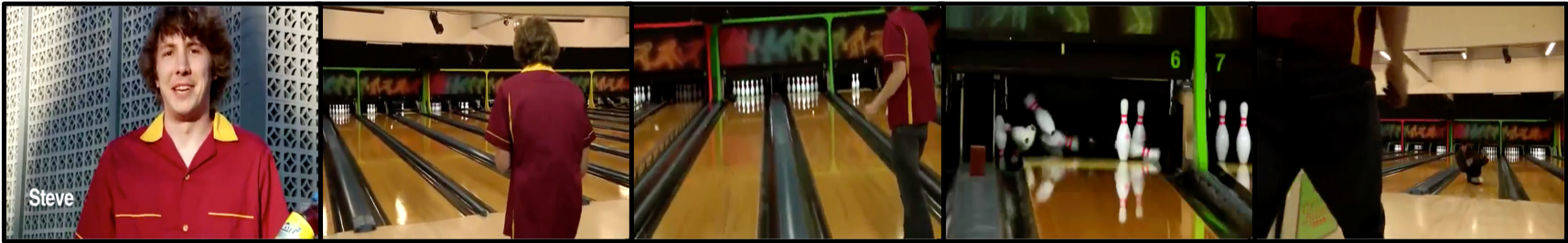}}} \\
		[5.5ex]
	\end{tabular}
	\caption{{\small \textbf{ActivityNet: Given Paragraph and Retrieve Video.} Negative Qualitative examples of \hse, \thse, and \fse on the task of given text to retrieve video. We mark the correct sample in \textbf{ \color{olive} green} and incorrect one in \textbf{ \color{red} red}.} }
	\label{anet_p2v_neg}
\end{figure}

\subsubsection{Qualitative Examples for \hse, \thse, and \fse on Retrieval Tasks.}
We show the qualitative examples on ActivityNet as below.
To show a systematic analysis of the success cases and failure cases, we choose to visualize positive examples of paragraph retrieval (in Figure~\ref{anet_v2p_pos}) and video retrieval (in Figure~\ref{anet_p2v_pos}) and negative examples in Figure~\ref{anet_v2p_neg} and Figure~\ref{anet_p2v_neg}.
We observe that in some failed cases, although \hse failed to retreive the correct text/video, it retreive very relevant item given the query information.

\end{document}